\documentclass[letterpaper, 10 pt, journal, twoside]{IEEEtran}

\IEEEoverridecommandlockouts                              

\usepackage{graphicx}
\usepackage{amsmath}
\usepackage{amssymb}
\usepackage{booktabs}
\usepackage{multirow}
\usepackage{rotating}
\usepackage{siunitx}
\usepackage[ruled,vlined,linesnumbered]{algorithm2e}
\usepackage[table,dvipsnames]{xcolor}
\usepackage{cuted}
\usepackage[inline]{enumitem}
\usepackage{microtype}
\usepackage{flushend}
\usepackage{cite}
\usepackage{bbm}
\usepackage[hang,flushmargin,symbol]{footmisc}

\def\eg{\emph{e.g}.}

\usepackage[colorlinks, breaklinks]{hyperref}

\usepackage[acronym]{glossaries}
\newacronym{rpn}{RPN}{Region Proposal Network}
\newacronym{roi}{RoI}{Region of Interest}
\newacronym{iou}{IoU}{intersection over union}
\newacronym{nms}{NMS}{non-maximum suppression}
\newacronym{rgcn}{R-GCNs}{Relational Graph Convolutional Networks~\cite{Schlichtkrull2018}}
\newacronym{fpn}{FPN}{Feature Proposal Network}
\newacronym{kern}{KERN}{Knowledge-Embedded Routing Network \cite{Chen2019}}
\newacronym{bert}{BERT}{Bidirectional Encoder Representations from Transformers~\cite{Devlin2019}}
\newacronym{gru}{GRU}{Gated Recurrent Cells~\cite{Li2016}}
\newacronym{fcn}{FCN}{fully-connected network}
\newacronym{rcgn}{RCGN}{Relational Graph Neural Network}
\newacronym{ggnn}{GGNN}{Gated-Graph Neural Network}
\newacronym{mlp}{MLP}{Multi-Layer Perceptron}
\newacronym{coco}{COCO 2017}{COCO 2017~\cite{lin2014microsoft}}
\newacronym{lvis}{LVIS v1}{LVIS v1~\cite{gupta2019lvis}}
\newacronym{svg}{SVG}{singular value decomposition}
\newacronym{kge}{KGE}{knowledge graph embedded}
\newacronym{sota}{SoTA}{state-of-the-art}
\newacronym{mlr}{MLR}{multi logistic regression}
\newacronym{vos}{VOS}{visualizing similarities between objects}
\newacronym{bdd}{BDD100k}{Berkeley Deep Drive~\cite{yu2020bdd100k}}
\newacronym{mot}{MOT}{Multi-Object Tracking}
\newacronym{mot17}{MOT17}{MOT17~\cite{MOT16}}
\newacronym{ssl}{SSL}{self-supervised learning}
\newacronym{fps}{FPS}{frames per second}
\newacronym{msa}{MSA}{minimum-sum assigment}
\newacronym{tempo}{TempO}{temporal ordering pretext}
\newacronym{nn}{NN}{nearest-neighbor}
\newacronym{ucf}{UCF101}{UCF101~\cite{soomro2012ucf101}}
\newacronym{hota}{HOTA}{higher-order tracking accuracy~\cite{luiten2020IJCV}}
\newacronym{rpd}{RPD}{random query patch detection~\cite{dai2021updetr}}
\glsdisablehyper

\usepackage[capitalize]{cleveref}
\crefname{section}{Sec.}{Secs.}
\Crefname{section}{Section}{Sections}
\Crefname{table}{Table}{Tables}
\crefname{table}{Tab.}{Tabs.}
\crefname{algorithm}{Algo.}{Algos.}

\newcommand{\figref}[1]{Fig.~\ref{#1}}
\newcommand{\tabref}[1]{Tab.~\ref{#1}}
\newcommand{\secref}[1]{Sec.~\ref{#1}}

\newcommand{\equref}[1]{Eq.~(\ref{#1})}

\renewcommand{\baselinestretch}{0.985}
\usepackage[resetlabels]{multibib}

\newcites{Suppl}{Supplementary References}

\newcommand{\change}[1]{\textcolor{black}{#1}}

\title{\LARGE \bf
Self-Supervised Representation Learning from Temporal Ordering of\\Automated Driving Sequences
}

\author{Christopher Lang$^{1,2}$, Alexander Braun$^{1}$,
    Lars Schillingmann$^{1}$,  Karsten Haug$^{1}$, and Abhinav Valada$^{2}$
    \thanks{$^{1}$Robert Bosch GmbH, Stuttgart, Germany}%
    \thanks{$^{2}$Department of Computer Science, University of Freiburg, Germany}
}

\begin{document}

\maketitle
\thispagestyle{empty}
\pagestyle{empty}

    \glsunsetall 
    \begin{abstract}
Self-supervised feature learning enables perception systems to benefit from the vast raw data recorded by vehicle fleets worldwide. 
While video-level self-supervised learning approaches have shown strong generalizability on classification tasks, the potential to learn dense representations from sequential data has been relatively unexplored. 
In this work, we propose TempO, a temporal ordering pretext task for pre-training region-level feature representations for perception tasks. 
We embed each frame by an unordered set of proposal feature vectors, a representation that is natural for object detection or tracking systems, and formulate the sequential ordering by predicting frame transition probabilities in a transformer-based multi-frame architecture whose complexity scales less than quadratic with respect to the sequence length. 
Extensive evaluations on the BDD100K, nuImages, and MOT17 datasets show that our TempO pre-training approach outperforms single-frame self-supervised learning methods as well as supervised transfer learning initialization strategies, achieving an improvement of $+0.7$ in mAP for object detection and $+2.0\%$ in the HOTA score for multi-object tracking.
\end{abstract}
       
    \section{Introduction}
    

Automated driving datasets typically cover only small sets of visual concepts, e.g., eight object classes in BDD100K~\cite{yu2020bdd100k} or ten classes in nuScenes~\cite{caesar2020nuscenes}, using a fixed camera setup. This narrow perspective can lead to poor transfer abilities, such that models struggle with driving scenes that are weakly represented in the training data. Consequently, very large datasets are required to cover the numerous traffic scenarios, which are expensive to curate and manually annotate. 
Self-supervised learning approaches offer a promising solution, as they allow network parameters to be pre-trained on pretext tasks that provide free supervision from consistency constraints in the raw data. The design of these pretext tasks should encourage the network to learn rich intermediate representations that emerge independent of human annotation. This has the potential to directly initialize model parameters by pre-training from in-domain and on-platform data and only fine-tune the network with limited annotated data on a downstream task.

\begin{figure}
    \centering
    \includegraphics[width=0.8\linewidth]{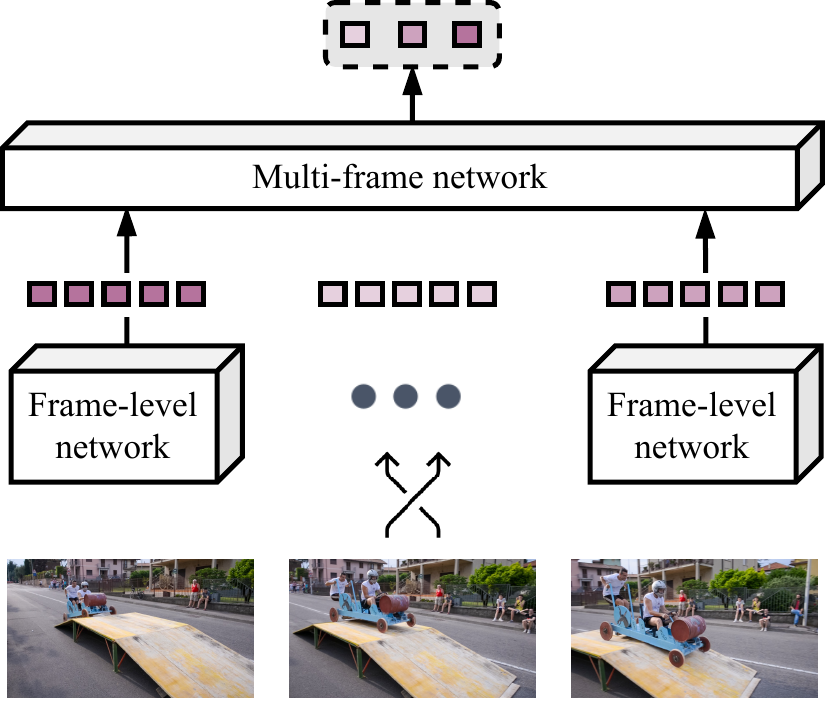}
    \caption{We introduce TempO, a self-supervised learning pretext task by the temporal ordering of frames, that pre-trains perception models for both frame-level tasks, such as object detection, and multi-frame tasks, such as multi-object tracking. We design a transformer-based multi-frame sorting architecture designed to scale quadrically with respect to sequence length.}
    \label{fig:intro_diagram}
\end{figure}

Self-supervised pre-training approaches have been demonstrated by image-level feature learning using contrastive methods~\cite{grill2020bootstrap,caron2020swav} that constrain the feature representations of an image to be invariant to a set of augmentations.
These methods were subsequently extended to region-level representation learning, where the contrastive approach is applied to image patches~\cite{dai2021updetr,xie2021pixpro}.
In robotic domains such as automated driving, the perception system observes the environment from a continuous stream of sensor data. 
Adding such temporal context to the pretext task allows learning from undistorted images by exploiting object permanence and dynamic constraints.
Recent methods enforce these by consistent appearance in tracking patches along a temporal cycle~\cite{wang2019learning,kong2020ccl, valverde2021there}.
Significantly, the task of ordering a set of shuffled frames into their original temporal succession~\cite{misra2016shuffle,lee2017unsupervised} has shown promising results, as it avoids restrictive assumptions and is never ill-defined. However, the existing formulations are limited to learning video-level feature representations by design. Those entangle information about an entire scene, which handicaps the transfer to dense prediction tasks such as object detection and tracking that require local predictions.\looseness=-1 

In this work, we hypothesize that the temporal ordering task has the potential for a rich understanding of local object semantics and interactions. 
Therefore, we propose TempO, a novel self-supervised pretext task for pre-training anchor-based object detection and tracking architectures based on the temporal ordering of frames.
Further, we state the temporal ordering task as a sequence estimation problem based on the tracking-by-detection paradigm, as depicted in \figref{fig:intro_diagram}. This breaks the combinatorial explosion in the computational complexity with respect to the sequence length in related works~\cite{xu2019self,misra2016shuffle} to be less than quadratic, which allows us to train on longer sequences with richer temporal context. We evaluate the performance of TempO pre-training on the BDD100K~\cite{yu2020bdd100k}, nuImages~\cite{caesar2020nuscenes}, and MOT17~\cite{MOT16} datasets by comparing them with pre-training on supervised datasets and existing self-supervised methods. 
\change{In addition, we conducted frame retrieval experiments to compare the pre-trained models with other related temporal pre-training methods without requiring a fine-tuning stage.}
We expect our results to be of significant interest to the robotics community, as they demonstrate that sequential data can be exploited for self-supervised pre-training of object detection and multi-object tracking models, thereby outperforming augmentation-based self-supervised pre-training.

The main contributions of this work are:
\begin{itemize}[noitemsep, topsep=0pt]
    \item We propose TempO, a self-supervised pre-training pipeline for object detection and multi-object tracking architectures.
    \item We redefine the temporal ordering as a frame-transition probability prediction and design a transformer-based sorting head that scales less than quadratic w.r.t. sequence length, enabling pre-training on longer sequences. 
    \item We present extensive ablation studies and evaluations of TempO pre-training for the downstream tasks of object detection and \gls{mot} that demonstrate the utility of our approach.
    \item We study the utility of representations learned from the self-supervised TempO pretext task for frame retrieval.
\end{itemize}

    \section{Related Work}
    {\parskip=3pt
\noindent\textit{Self-Supervised Image Feature Learning}: Self-supervised learning has been studied extensively on single images, where the field can be broadly categorized into image-level, region-level, and pixel-level approaches.
Image-level approaches learn a global embedding vector per frame and are typically evaluated on image classification tasks~\cite{chen2020improved,caron2021dino}. 
Instance-based discrimination uses a contrastive loss function that minimizes similarity metrics among different views of the same image (positives) while maximizing similarity for various other images (negatives). 
Clustering-based methods~\cite{caron2020swav} circumvent the negative image selection by using cluster assignments of augmented views as a supervision signal. 
The temporal consistency in video clips is also used as a source for positive pairs from subsequent frames~\cite{feichtenhofer2021large,xu2021vfs,ma2022vip}. \change{Such methods learn features from unaugmented data, and the authors observed object-level correspondences emerging in the learned feature representations~\cite{xu2021vfs}.}

Nevertheless, many computer vision tasks, including object detection~\cite{lang2022robust, lang2022hyperbolic} and segmentation~\cite{gosala2022bird, mohan2022amodal}, require dense representations that also encode local image information. Region-level approaches~\cite{dai2021updetr,Ding2022DeeplyUP}, therefore, rely on pretext tasks that operate on patches of an image or the feature map. 
Patch discrimination methods~\cite{o2020vader,liu2020selfemd} utilize these image patches to learn local augmentation-invariant embeddings analogously to the image-level methods described above. The patch re-identification pretext task~\cite{ding2022dupr,dai2021updetr,xie2021pixpro,Xie2021DetCoUC}, on the other hand, combine patch discrimination and re-identification pretext tasks for object detection pretraining, as the losses include regression and classification-based terms analogously to the downstream task.
In our experiments, we compare against UP-DETR~\cite{dai2021updetr} as a baseline, which is trained on localizing random crops in an image and contrasting the feature embeddings of crops within an image. 

In recent years, sequential data has been explored as a supervisory signal to learn dense feature representations. 
Tracking approaches predict the offset of an image region in feature space forward and backward along a cycle in time~\cite{wang2019learning,jabri2020space}, whereby the difference between start and end points is used as a training signal. However, such methods operate only on pairwise frame contexts and depend on data domains where the presence of an object over a certain time window can be ensured.\looseness=-1

\begin{figure*}
    \centering
    \includegraphics[width=0.9\textwidth]{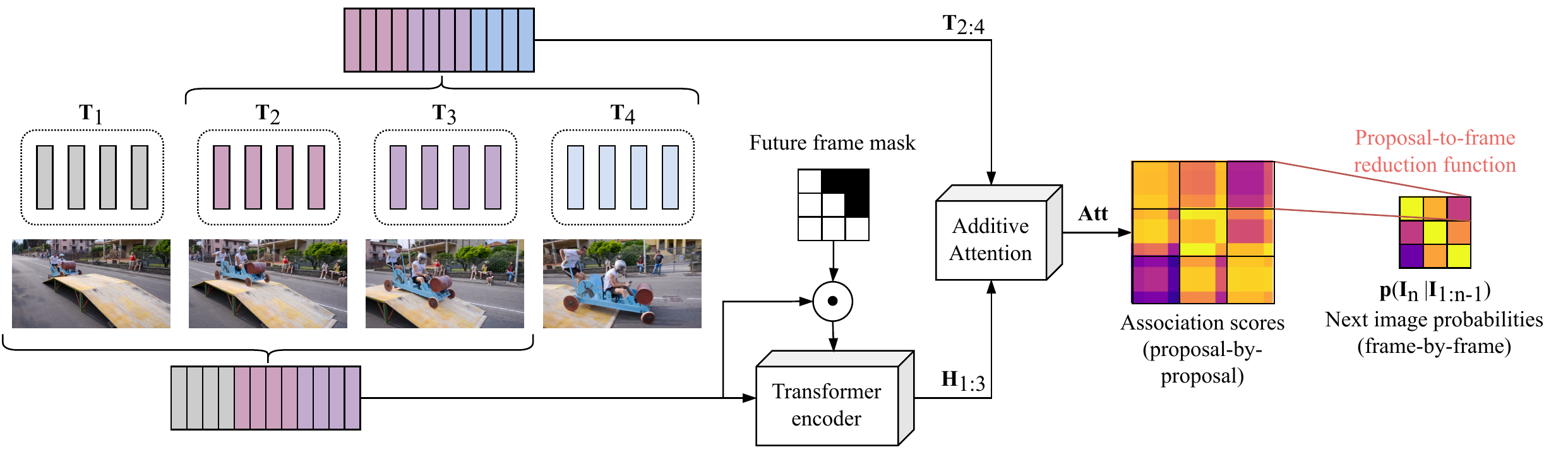}
    \caption{Illustration of the multi-frame head in the TempO architecture. Each frame $n$ is represented by a set of proposal feature vectors $\mathbf{T}_n$ extracted by the same frame-level network. The proposal features are concatenated and encoded by a transformer encoder into a set of history tokens $\mathbf{H}$ using a future frame masking that allows each proposal to aggregate temporal context from past frames only.
    An additive attention mechanism then computes the association scores $\mathbf{Att}$ between the history tokens and the proposal feature vectors. We next map all scores of proposal features corresponding to the same frame onto scalar image transition probabilities. 
    Our proposed temporal ordering task maximizes these probabilities for the correct temporal order during pre-training. 
    }
    \label{fig:loss_architecture}
\end{figure*}

{\parskip=3pt
\noindent\textit{Self-Supervised Video Feature Learning}:
Video feature learning exploits the temporal consistency in sequential data as a supervision signal to generate video- \change{or frame-}level embeddings. Self-supervised learning approaches can be broadly categorized into three types of pretext tasks:
\begin{enumerate*}
    \item Contrastive methods learn video~\cite{qian2021spatiotemporal,Ding2022DeeplyUP} invariant to a set of temporally consistent augmentations.
    \item Sequential verification performs a binary classification if a sequence is correct or in a shuffled temporal order~\cite{misra2016shuffle}. Other formulations discriminate between forward and backward order of frames~\cite{wei2018learning}. 
    Such methods are used for encoding video clips all at once but are outperformed by temporal ordering methods as the network needs to reason about richer structure in image sequences
    \item Temporal ordering methods on video classification tasks~\cite{lee2017unsupervised,wei2018learning} is formulated as a classification problem among all the frame permutations, which limits their usage due to the combinatorial explosion of orderings.
\end{enumerate*}
Since the aforementioned methods learn video-level embeddings, they are only evaluated on frame retrieval and action recognition tasks, \change{which is why we evaluate our method against them on the frame-retrieval task in \secref{sec:frame_retrivial}}.
In addition, we define the temporal ordering pretext task for learning frame-level feature representations that allow for a larger variety of downstream tasks, as described in the following section. 
}

    \section{Technical Approach}
This section details our proposed TempO pretext task for self-supervised learning of region-level visual feature representations from a temporal context. 
We first introduce the self-supervised pretext task in \secref{sec:ordering_task}, followed by the network architecture in \secref{sec:singe_frame_network} and \secref{sec:ordering_network}, and finally describe the transfer learning technique to perform downstream task evaluations in \secref{sec:downstream_networks}.

\subsection{Sequence Ordering Task Definition}
\label{sec:ordering_task}

Our proposed TempO pretext task requires image sequences of length $N_{seq}$. 
It defines a training sample by using the first frame in the sequence $\mathcal{I}_1$ as an anchor in time and the remaining sequence frames $\mathcal{I}_{2:N}$ in arbitrary order. In the first step, a single-frame network extracts an unordered set of proposal feature vectors, treating each frame independently.
The multi-frame transformer head then processes the concatenated proposal feature vectors over all $N_{seq}$ frames in a training sample and maps them onto next-image probabilities given a sequence of images as described in \secref{sec:ordering_network}. 
Our training objective is to maximize the next-image probability $\rho \left(\mathcal{I}_n |  \mathcal{I}_{1 : n-1}\right)$ for the observed temporal orderings in the video data using a ranking loss formulation

{\small
\begin{equation}\label{eq:tempo}
    \mathcal{L} = \sum_{m \neq n} \max \left\{ \rho(\mathcal{I}_m | \mathcal{I}_{1 : n-1} ) -  \rho(\mathcal{I}_n | \mathcal{I}_{1 : n-1}) + \Delta , 0\right\},
\end{equation}
}%
where $\Delta \geq 0$ is a scalar margin.


\subsection{Single-Frame Network}
\label{sec:singe_frame_network}

Our approach adapts to network architectures that process images and yield a set of $P$ proposal feature vectors $\mathbf{Q} \in \mathbb{R}^{P \times D}$ of dimension $D$ per frame. 
This includes common region proposal-based~\cite{peize2020sparse} and transformer-based~\cite{zhu2020deformable} architectures. 
The majority of our experiments are conducted on the Sparse R-CNN~\cite{peize2020sparse} object detection architecture, using a ResNet-50~\cite{he2016resnet} with a feature pyramid network as a feature extractor. 
It learns a sparse set of $P$ proposal boxes and features from which classification scores and bounding boxes are generated. The initial proposal features are extracted from learned proposal box regions in the feature map. The proposal features are then iteratively refined by a sequence of dynamic heads, enabling interaction between proposal features via self-attention.\looseness=-1

We implement two distinct branches of dynamic heads: a detection branch that extracts object proposal features, consisting of six iterative heads, and a tracking branch that extracts tracking proposal features from two iterative dynamic heads that are used to associate object identities throughout a sequence. 
The model under pre-training uses two dynamic heads, whose parameters are cloned to initialize the parameters of both the detection
and tracking branch (see \secref{sec:ordering_network}) during fine-tuning. The motivation for separating the tracking and detection branches is that the feature representations pursue competing objectives. While detection features should learn to generalize across an object type (\eg car), tracking features should learn to discriminate between object instances.

\subsection{Multi-Frame Sequence Ordering Network}
\label{sec:ordering_network}

The overall setup of our sequence ordering network is depicted in~\figref{fig:loss_architecture}. 
We express the image transition probability $\rho\left(\mathcal{I}_n |  \mathcal{I}_{1 : n-1}\right) = \rho(\mathbf{T}_n | \mathbf{H}_{n-1})$ by \change{computing a similarity score between the track features $\mathbf{T}_{n}$ for frame $n$ and a sequence of history tokens  $\mathbf{H}_{1:n-1} \in \mathbb{R}^{(P \times D})$}.
The \change{history tokens at $\mathbf{H}_{n-1}$ are computed as the output of a transformer encoder $\mathbf{H}_{n-1} = f\left( \mathbf{T}_{1}, \dots, \mathbf{T}_{n-1}\right)$} that takes as input the track feature vectors up to frame ${n-1}$. The temporal succession is implemented by masking track features of future frames in the attention matrix.
Finally, we compute additive attention weights between the history tokens and the track features, which yields the association score matrix $\mathbf{Att}$.

{\parskip=3pt
\noindent\textit{Proposal-to-Frame Reduction Function}:
The final stage in our multi-frame head is the reduction from an association score matrix $\mathbf{Att}$ to a next-frame transition probability. 
As the associations are derived from unordered sets of proposals, this function must be permutation invariant with respect to the elements of the association matrix. We, therefore, employ the average of all matrix elements (AvgPool) that encourage similarity between all track features in temporally subsequent frames.}

\subsection{Downstream Task Architectures}
\label{sec:downstream_networks}

We use the Sparse R-CNN~\cite{peize2020sparse} architecture as our base model. Our variant builds on ResNet-50~\cite{he2016resnet} as the backbone and two iterative dynamic heads for generating 100 object proposals per frame. 
We collect feature vectors over all sequence frames for the pretext task and feed them to a masked transformer encoder described in \secref{sec:ordering_network}.

{\parskip=3pt
\noindent\textit{Object Detection}:
For the object detection fine-tuning, we adapt the originally proposed Sparse R-CNN~\cite{peize2020sparse}. Therefore, we build upon the pre-trained Sparse R-CNN architecture and stack four iterative dynamic heads, called the detection branch, on top of the pre-trained heads. The final object proposal vectors are mapped onto classification scores and bounding box regression coordinates using separate linear layers.
}
 
{\parskip=3pt
\noindent\textit{Multi-Object Tracking}:
For \gls{mot}, we associate proposals based on their additive attention between track features $\mathbf{T}_n$ and the history features of the previous frame $\mathbf{H}_{n-1}$ that is generated by the transformer encoder. We follow the setup in QDTrack~\cite{pang2021quasi}, employing their bidirectional softmax matching in feature space, tracker logic, and training pipeline. We further extend the model into an object detector, as described above. We clone the pre-trained iterative heads to have distinct parameter sets for the tracking and detection branches.
}

    \section{Experiments Evaluation}
    \glsunsetall

We pre-train the networks using our proposed TempO approach on the \gls{bdd}, nuImages, and \gls{mot17} datasets.
In this section, we evaluate the performance of our TempO pre-trained models for single-frame and multi-frame downstream perception tasks. 
\change{Additionally, we benchmark the performance against other standard initialization strategies for object detectors, as well as related video-based pre-training methods in \secref{sec:downstream_benchmarks}. 
Finally, we ablate over key aspects of our pre-text design in \secref{sec:ablation_study}, and we benchmark the representations learned by our proposed method on the frame retrieval task without fine-tuning in \secref{sec:frame_retrivial}. In the supplementary material, we complement our analysis by studying the convergence behavior in \secref{sec:convergence} and the influence of camera ego-motion on the tracking performance in \secref{sec:egomotion}.}


\subsection{Datasets}
\label{sec:datasets}

The BDD100k dataset contains 100,000 crowdsourced videos of driving scenes such as city streets and highways. 
\Gls{mot} annotations are available for a subset of 1400 videos at 5 \gls{fps} and cover eight object categories of overall 131,000 identities. For our pre-training, we sample frames at 5 \gls{fps} from all videos, excluding videos in the val and test sets. \change{From each sequence, we generate training samples for non-overlapping intervals of $N_{seq}$ frames.}
\change{
The nuImages dataset is part of the nuScenes~\cite{caesar2020nuscenes} ecosystem, providing 60k train, 16k val, and 16k test images with 2D bounding box annotations. The dataset provides six past and six future frames for each annotated image. For \textit{TempO} pre-training, we generate one training sample from each of these sequences in the train set, with the annotated image at its center.
}
The MOT17 challenge consists of 14 video sequences (7 training and 7 testing) at varying frame rates ($>$14fps) and sequence lengths ($>$20s) in unconstrained environments filmed with both static and moving cameras. It provides \gls{mot} annotations that feature a people class. The training videos down-sampled to  5 \gls{fps} are added to the \textit{TempO} pre-training set.

\begin{figure*}[t]
    \centering
    \footnotesize
    \begin{minipage}[t]{0.49\textwidth}
        \centering
        \begin{minipage}{0.48\textwidth}
            \includegraphics[trim={6cm 0 0 0},clip,width=\textwidth]{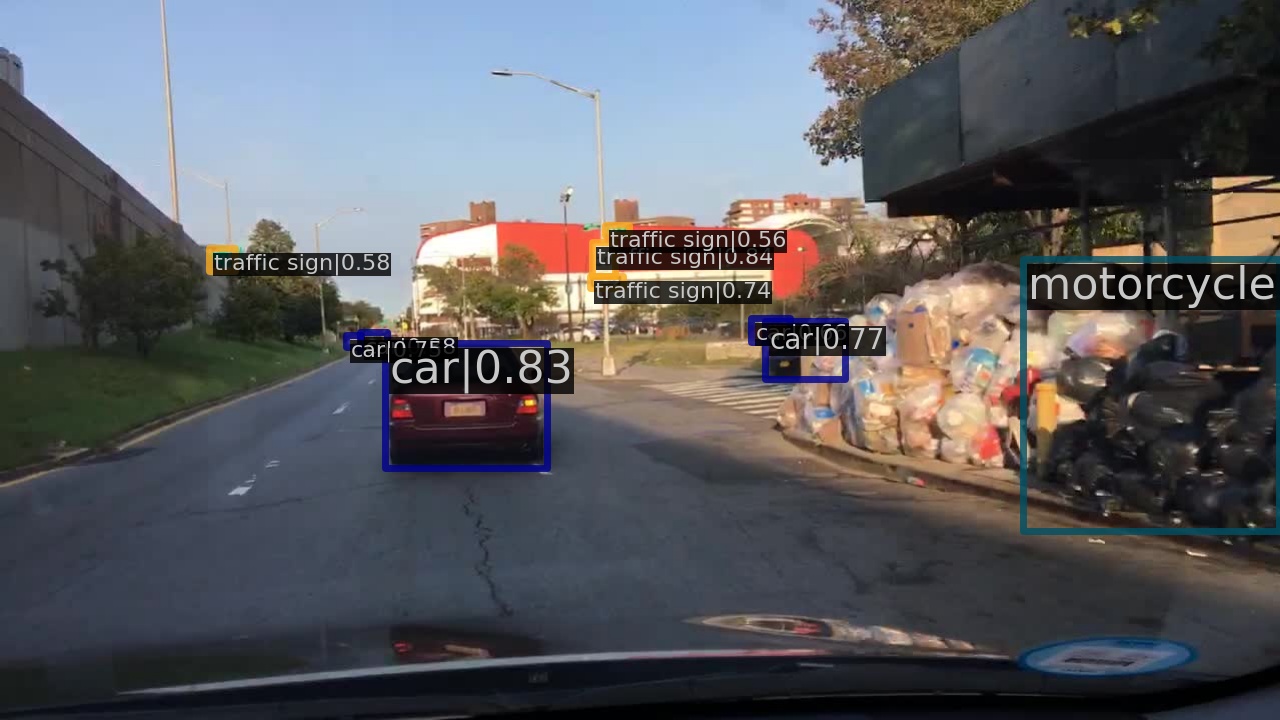}
        \end{minipage}%
        \hspace{0.01\textwidth}%
        \begin{minipage}{0.48\textwidth}\centering
            \includegraphics[trim={0 0 6cm 0},clip,width=\textwidth]{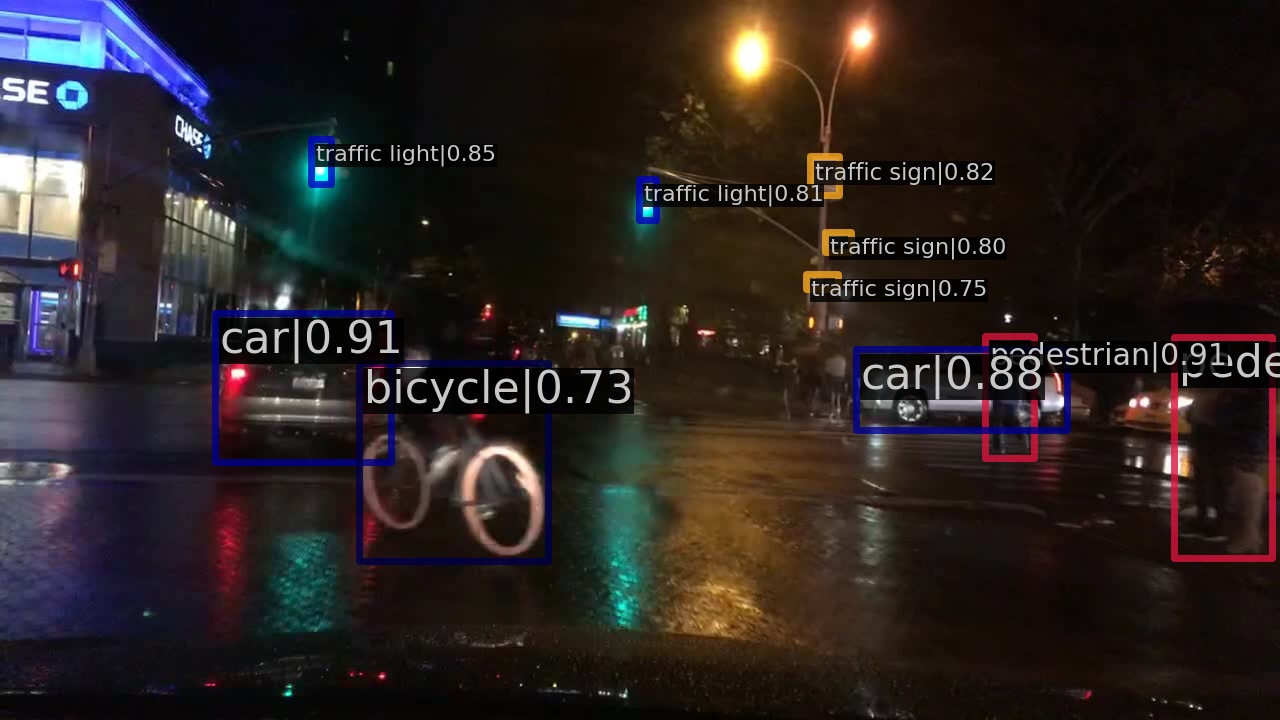}
        \end{minipage}\\[1mm]{a) ImageNet initialized, trained for 12 epochs on \gls{bdd}}
    \end{minipage}
    \hfill
    \begin{minipage}[t]{0.49\textwidth}
        \centering
        \begin{minipage}{0.48\textwidth}
            \includegraphics[trim={6cm 0 0 0},clip,width=\textwidth]{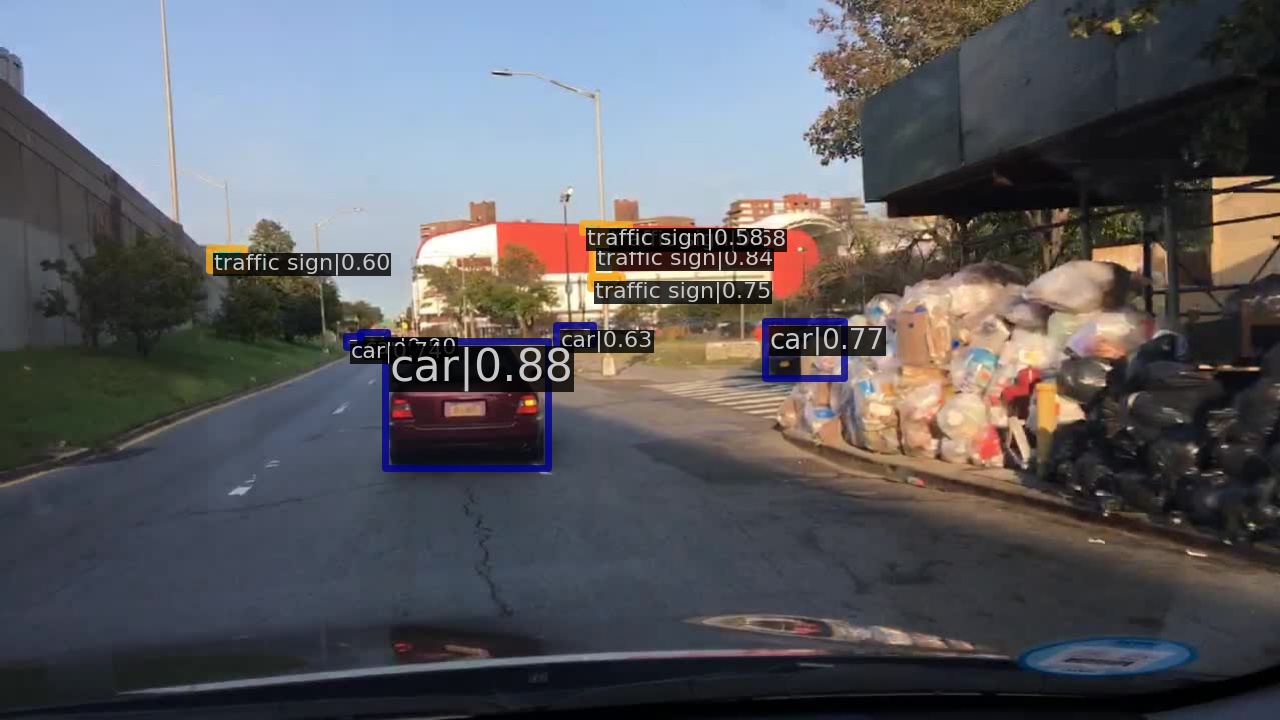}
        \end{minipage}%
        \hspace{0.01\textwidth}%
        \begin{minipage}{0.48\textwidth}\centering
            \includegraphics[trim={0 0 6cm 0},clip,width=\textwidth]{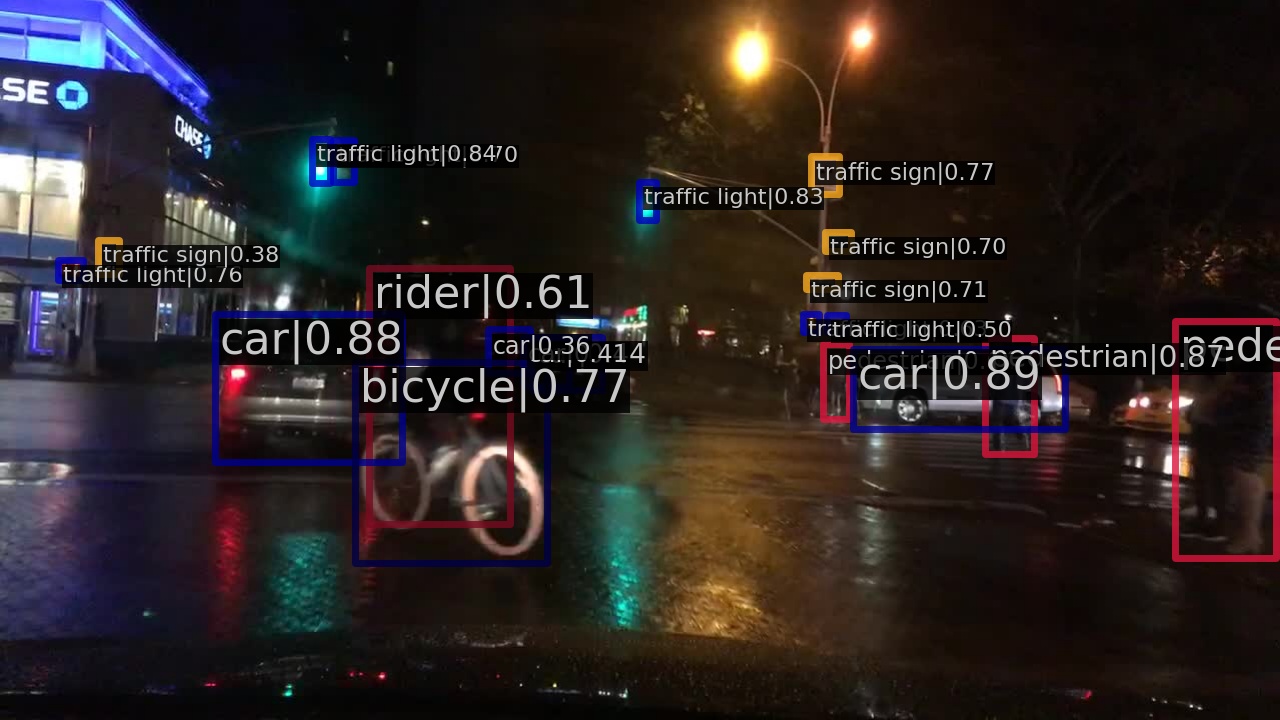}
        \end{minipage}\\[1mm]{b) \textit{TempO} pre-trained, fine-tuned for 6 epochs on \gls{bdd}.}
    \end{minipage}

    \caption{Comparison of qualitative object detection results on the \gls{bdd} val set using the Sparse R-CNN detector with standard (a) and \textit{TempO} (b) parameter initialization strategies. 
    Observe that the \textit{TempO} pre-trained detector avoids a ghost detection of a motorcycle within the garbage bags and detects the poorly lit rider on top of the moving bicycle. \change{Please refer to the supplementary material for a comparison of both 6-epochs and 12-epochs fine-tuned models.}}
    \label{fig:qualitative_example_od}
\end{figure*}


\subsection{Training Protocol}
\label{sec:trainingprotocol}

We pre-train the models for six epochs on four \change{NVIDIA V100 GPUs} with a batch size of 8. Thereby, we \change{normalize each color channel separately}. We use the AdamW optimizer with an initial learning rate of $2.5 \cdot 10^{-5}$ and weight decay of $10^{-4}$. A step scheduler further reduces the learning rate by a factor of 10 every three epochs.
We fine-tune the models for another six epochs on the respective downstream tasks. For fine-tuning, we resize the images to a resolution of 800 pixels on the longer side and perform spatial augmentations, including random cropping or flipping, and photometric augmentations, including brightness and hue variation. \change{Please refer to \secref{sec:implementation} of the supplementary material for additional details on hyperparameter tuning strategies and computational requirements.}


\subsection{Evaluation of Pre-training Methods on Downstream Tasks}
\label{sec:downstream_benchmarks}

We evaluate how the region-level feature representations learned from the TempO pre-training impact the performance of single-frame and multi-frame downstream tasks, i.e., for object detection \change{on the BDD100k and nuImages~\cite{caesar2020nuscenes}} \change{using average precision metrics}, and multi-object tracking on the BDD100k as well as the MOT17 dataset \change{using CLEAR and HOTA~\cite{luiten2020hota} metrics}.
We follow the evaluation protocol as in~\cite{xu2019self,caron2021dino,chen2020improved}, where we fine-tune all models for the same fixed number of epochs (12 in our case). Our TempO models were pre-trained on ordering sequences of length $N=8$, using two layers in the transformer encoder for six epochs. 

\subsubsection{Object Detection Results}

\tabref{tab:eval_det_bdd100k_test} shows the mean average precision over all the classes in the \gls{bdd} object detection benchmark. 
We compare our proposed self-supervised pre-training approach against various initialization strategies using the Sparse R-CNN~\cite{peize2020sparse} object detector, including the common practice of pre-training the feature extractor as a classifier on the ImageNet dataset (referred to as Cls), as well as pre-training the model parameters on the supervised COCO 2017~\cite{lin2014microsoft} object detection dataset (Det). Additionally, we compare against the single-frame self-supervised pre-training task of random query patch detection (RPD)\cite{dai2021updetr}. 

\begin{table}[t]
\footnotesize
\centering
\setlength{\tabcolsep}{5pt}
\caption{Comparison of parameter initialization strategies for the downstream task of object detection on the BDD100k val set. 
\textit{TempO} pre-training uses a sequence length of 8 frames, two layers in the multi-frame network, and AvgPool.
}
\label{tab:eval_det_bdd100k_test}
\begin{tabular}{lll|p{0.35cm}p{0.35cm}p{0.35cm}p{0.35cm}} \toprule
Model          & \multicolumn{2}{c}{\change{Pre-train}}       & \multicolumn{4}{c}{BDD100k val detection}              \\
               & Dataset           & Task                 & AP            & APs           & APm           & APl           \\ \midrule
Faster R-CNN   & ImageNet          & Cls                   & 23.1          & 10.0          & 28.0          & 38.8          \\
FCOS           &                   & Cls & 25.9          & 7.6           & 31.2          & 52.0          \\
Sparse R-CNN   &                   & Cls & 24.9          & 10.7          & 31.2          & 42.4          \\
Sparse R-CNN   & COCO              & Det     & 30.7          & 15.2          & 34.3          & 50.8          \\
DeformableDETR &                   & Det     & 30.2          & 14.2          & 34.0          & 51.3          \\ 
Sparse R-CNN   & BDD100k           & RPD                  & 29.4          & 14.3          & 32.8          & 49.1          \\ \midrule
Faster R-CNN   & BDD100k           & TempO                & 24.3          & 12.6          & 30.3          & 42.1          \\
DeformableDETR &                   & TempO                & 30.8          & 14.5          & 34.9          & \textbf{53.8} \\
Sparse R-CNN   &                   & TempO                & \textbf{31.4} & \textbf{15.7} & \textbf{35.2} & 51.6   \\ \bottomrule      
\end{tabular}
\end{table}

\begin{table}[t]
\footnotesize
\centering
\caption{\change{Comparison of different parameter initialization strategies for object detection on the nuImages dataset. 
}}
\label{tab:eval_det_nuimages}
\begin{tabular}{lllllll} \toprule
Model        & \multicolumn{2}{c}{\change{Pre-train}} & \multicolumn{4}{c}{nuImages val detection} \\
             & Dataset         & Task        & AP        & APs      & APm      & APl      \\ \midrule
Faster R-CNN & ImageNet        & Cls         & 29.8      & 16.9     & 28.3     & 40.9     \\
Faster R-CNN & COCO            & Det         & 31.6      & 17.0     & 30.4     & 42.2     \\
Sparse R-CNN & COCO            & Det         & 30.3      & 15.8     & 30.1     & 40.5     \\ \midrule
Sparse R-CNN & BDD100k         & TempO       & 30.9      &  17.1        &  29.3        &   42.1       \\
Faster R-CNN & nuScenes        & TempO       & 34.3      & 18.3     & 34.0     & 42.1     \\
Sparse R-CNN & nuScenes        & TempO       & \textbf{34.6}      & \textbf{18.4}     & \textbf{34.4}     & \textbf{46.6}     \\ \bottomrule
\end{tabular}
\end{table}

We observe that the TempO initialization outperforms supervised pre-training on the COCO 2017 dataset by $+0.7\%$ in mAP, while the performance gain compared to the single-frame pretext task of random query patch detection is as large as $+2\%$ in mAP. 
Moreover, the convergence experiments in \secref{sec:convergence} of the supplementary material show that TempO pre-training results in faster convergence of the detectors. 
In \figref{fig:qualitative_example_od}, we present qualitative results of (a) a Sparse R-CNN object detector trained with ImageNet pre-trained weights and (b) the TempO pre-trained Sparse R-CNN detector. 
The TempO pre-trained Sparse R-CNN detector improves detection quality by suppressing a ghost detection of a motorcycle within the pile of garbage bags on the right side of the image shown in \figref{fig:qualitative_example_od}~(b). It also detects the poorly illuminated rider on the moving bicycle in \figref{fig:qualitative_example_od}~(b). 
\change{These behaviors indicate a broader understanding of object relations and appearance variations that emerge from a more extensive training corpus and an open-world training objective.}

\change{We observe similar performance benefits for pre-training and fine-tuning on the nuImages benchmark in \tabref{tab:eval_det_nuimages}. The TempO pre-trained parameter initialization consistently outperforms its supervised counterparts for COCO 2017 and ImageNet pre-trained models by $\geq + 2.7\%$ in mAP. Please note that the TempO pre-training on the BDD100k train set outperforms the COCO pre-trained initialization by $+ 0.6\%$ in mAP.}

{\parskip=3pt
\noindent\textit{Comparison to Scene-level Self-supervised Learning Methods}: 
\label{sec:downstream_video_benchmark}
In \tabref{tab:rebuttal}, we compare \textit{TempO} pre-training with self-supervised pre-training methods that learn only scene-level feature descriptors, such as scene-level feature learning strategies from frame-based~\cite{caron2021dino,chen2020improved} and sequence-based~\cite{xu2019self,misra2016shuffle,xu2021vfs} pretext tasks.

\begin{table}[t]
\footnotesize
\centering
\setlength{\tabcolsep}{5pt}
\caption{Comparison of different self-supervised parameter initialization strategies for object detection using the Sparse R-CNN model with ResNet-50 backbone on the BDD100k val set. 
\change{\textit{Det} refers to object detection, \textit{Shuffle} to the frame order verification~\cite{misra2016shuffle}, and \textit{Order} to frame order classification~\cite{xu2019self}}.
}
\label{tab:rebuttal}
\begin{tabular}{ll|p{0.35cm}p{0.35cm}p{0.35cm}p{0.35cm}p{0.35cm}p{0.35cm}} \toprule
\multicolumn{2}{c}{\change{Pre-train}} & \multicolumn{6}{c}{BDD100k val object detection} \\ 
Dataset & Task        & AP   & AP50 & AP75 & APs  & APm  & APl  \\ \midrule
COCO    & Det & \underline{27.5} & 49.9 & \underline{25.8} & 12.6 & 30.5 & \underline{47.8} \\
BDD100k & MoCo v2~\cite{chen2020improved} & 24.8 & 46.7 & 22.4 & 12.7 & 28.3 & 39.1 \\
        & DINO~\cite{caron2021dino}       & 26.7 & 50.0 & 24.3 & 13.0 & 29.9 & 46.1 \\
        & Shuffle~\cite{misra2016shuffle} & 27.4 & \underline{51.1} & 25.0 & \underline{13.8} & \underline{31.3} & 43.5 \\
        & Order~\cite{xu2019self}         & 27.0 & 49.7 & 24.8 & 12.3 & 30.5 & 46.2 \\
        & \change{VFS~\cite{xu2021vfs}} & 26.9 & 46.6 & 26.1 & 11.9 & 31.9 & 47.2 \\
        \midrule
BDD100k & TempO (Ours)  & \textbf{30.7} & \textbf{55.5} & \textbf{28.6} & \textbf{15.4} & \textbf{34.6} & \textbf{49.6} \\ \bottomrule
\end{tabular}
\end{table}

For the main results, we found such methods only partially comparable with our approach since the parameters of the object detection architecture's neck and head are not initialized with the SSL pre-trained weights. In the following, we therefore randomly initialize parameters of the detection architecture's neck and head. We trained the single-frame methods on double bounding box crops from the \acrshort{bdd} detection train set and the sequence-based methods on the same dataset as our TempO method, using the training protocol described in \secref{sec:trainingprotocol}. For frame order verification~\cite{misra2016shuffle} and classification~\cite{xu2019self}, we used a sequence length of 3 and 5, respectively. 
\change{For video frame-level similarity~\cite{xu2021vfs} pre-training, we choose a continuous sampling strategy with an interval of 4 to span the same time period as the TempO approach. We train VFS without negative pairs, as it was reported to achieve higher object tracking accuracy.}

Our TempO approach outperforms \change{temporal ordering and temporal contrastive self-supervised learning methods} by $\geq +3.3\%$ in mAP on the BDD100k validation set. 
This adds to the main experimental results, which showed that our proposed approach also outperforms initialization strategies that pre-train all detector parameters, \eg~from supervised training on the COCO 2017 dataset or self-supervised training on re-localization of random image patches (RPD). 
}

\subsubsection{Multi-Object Tracking Results}

\begin{figure*}
    \centering
    \begin{minipage}[b]{0.15\textwidth}
        \centering
        \includegraphics[trim=0 110 550 60, clip, width=\textwidth]{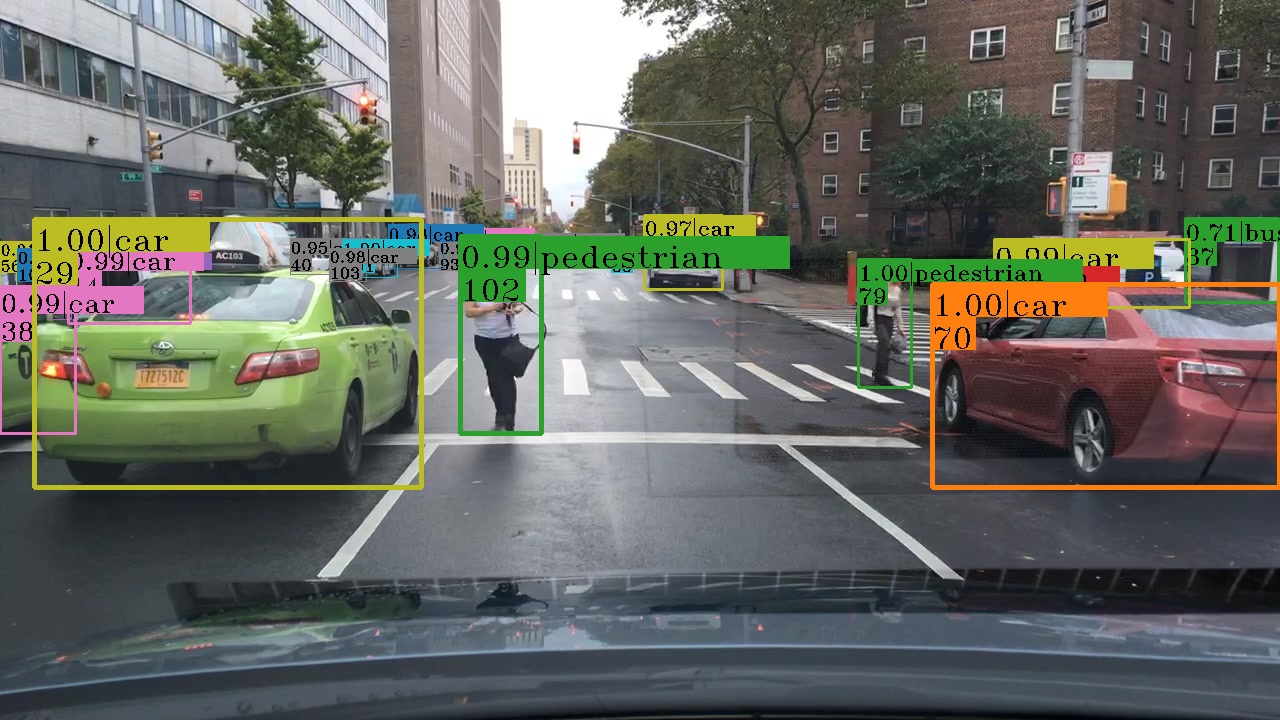}\\[1pt]{\scriptsize\textit{a)} Frame 85}
    \end{minipage}
    \hfill
    \begin{minipage}[b]{0.15\textwidth}
        \centering
        \includegraphics[trim=0 110 550 60, clip, width=\textwidth]{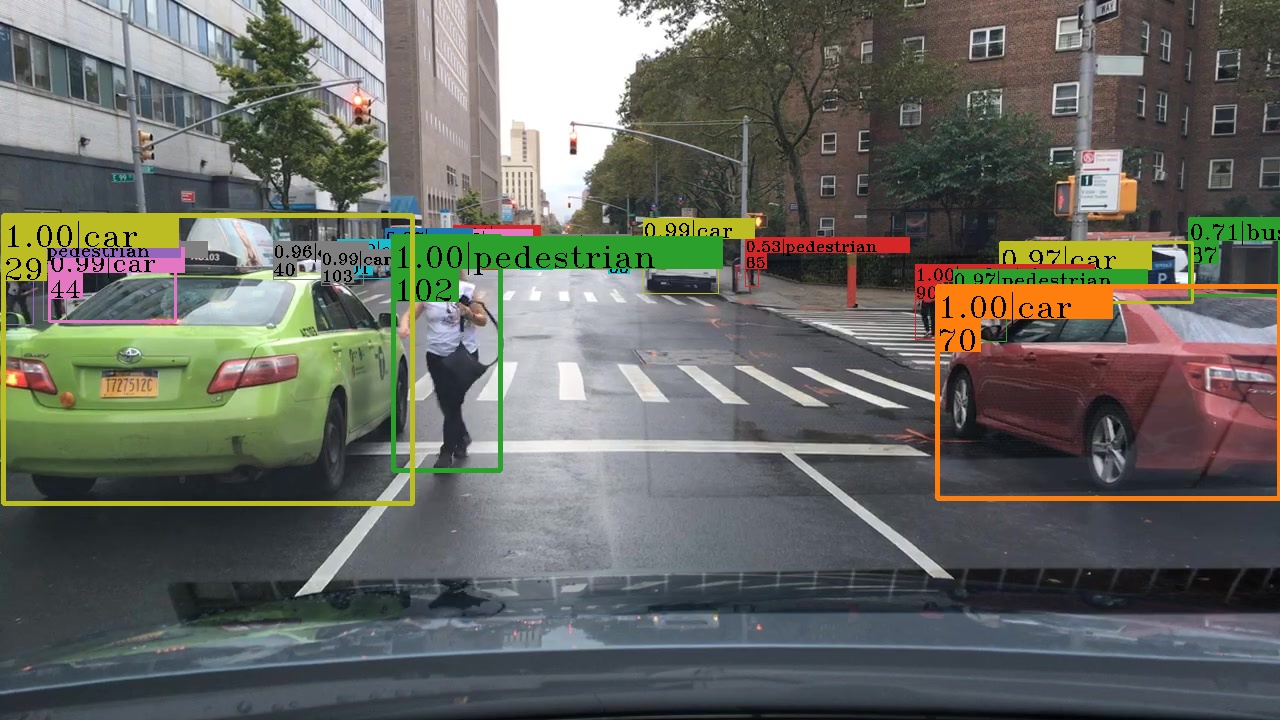}\\[1pt]{\scriptsize\textit{b)} Frame 90}
    \end{minipage}
    \hfill
    \begin{minipage}[b]{0.15\textwidth}
        \centering
        \includegraphics[trim=0 110 550 60, clip, width=\textwidth]{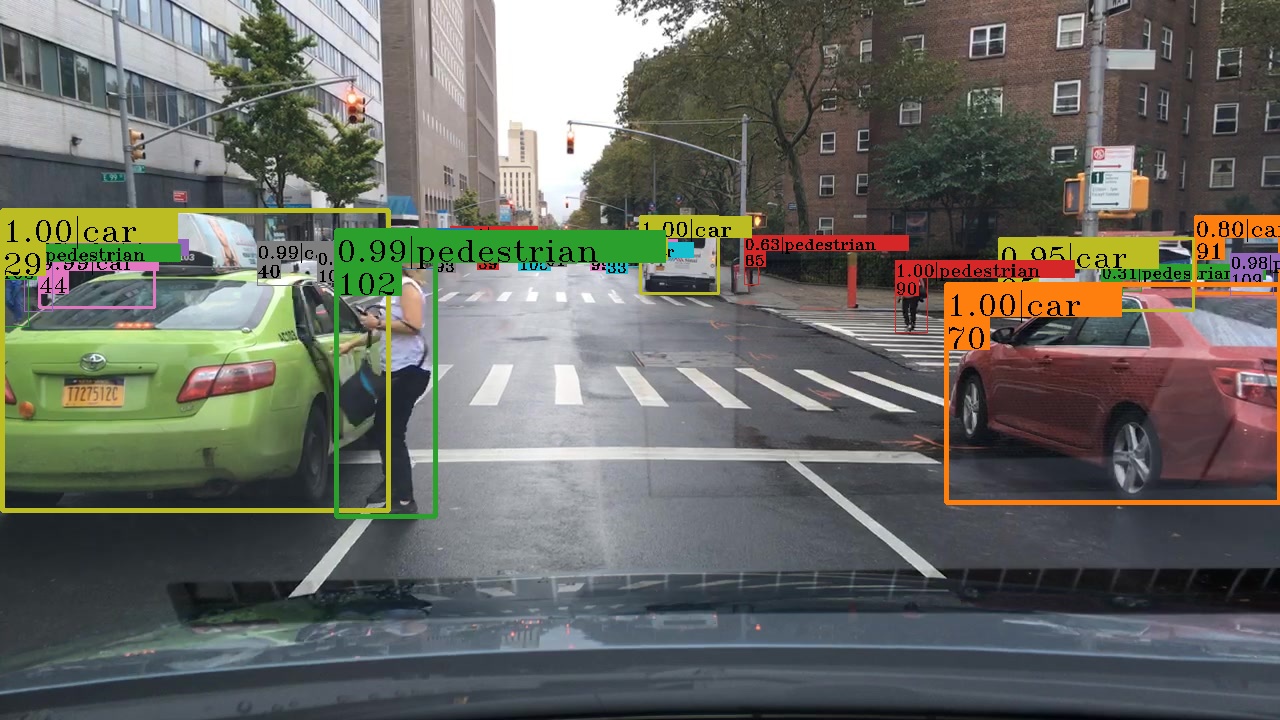}\\[1pt]{\scriptsize\textit{c)} Frame 95}
    \end{minipage}
    \hfill
    \begin{minipage}[b]{0.15\textwidth}
        \centering
        \includegraphics[trim=0 110 550 60, clip, width=\textwidth]{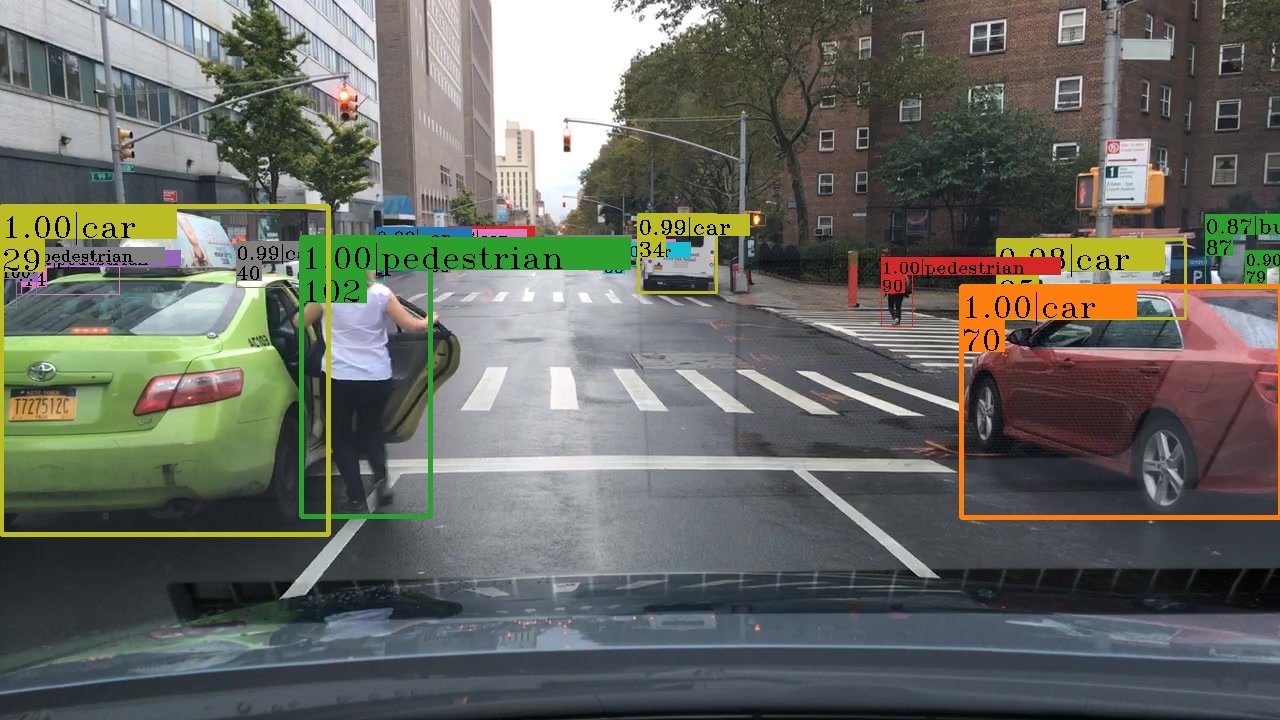}\\[1pt]{\scriptsize\textit{d)} Frame 100}
    \end{minipage}
    \hfill
    \begin{minipage}[b]{0.15\textwidth}
        \centering
        \includegraphics[trim=0 110 550 60, clip, width=\textwidth]{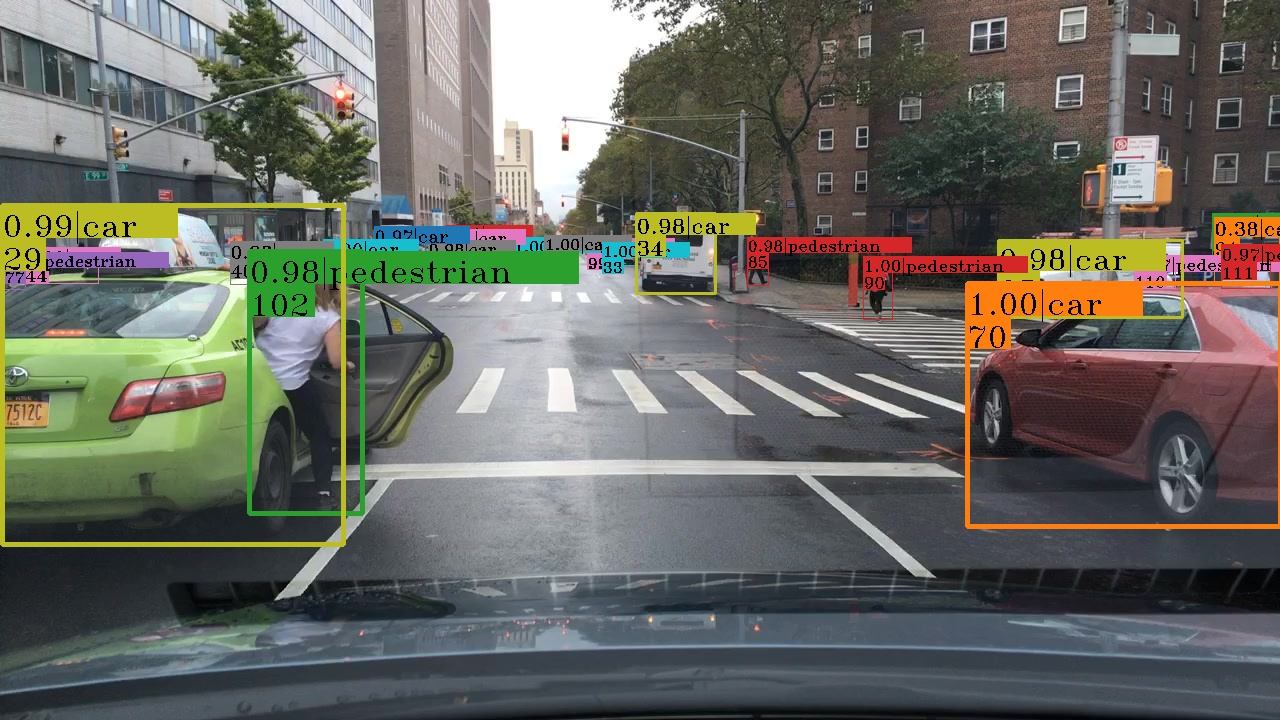}\\[1pt]{\scriptsize\textit{e)} Frame 105}
    \end{minipage}
    \hfill
    \begin{minipage}[b]{0.15\textwidth}
        \centering
        \includegraphics[trim=0 110 550 60, clip, width=\textwidth]{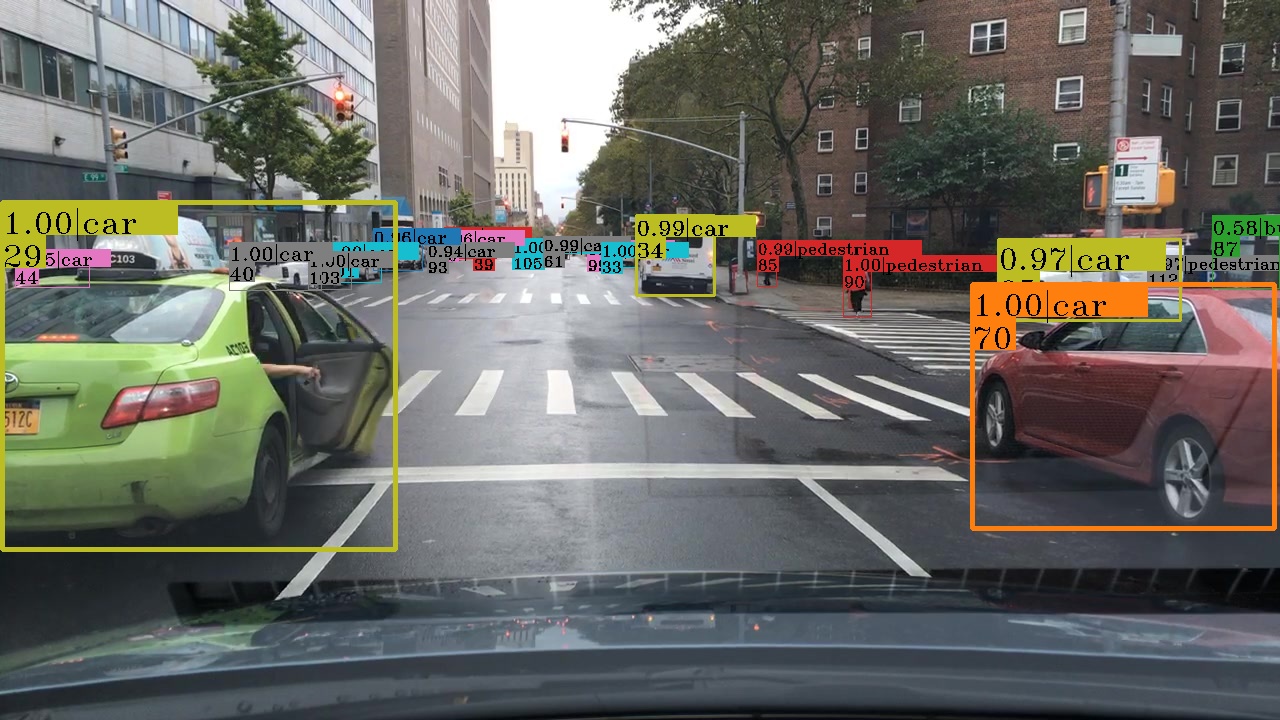}\\[1pt]{\scriptsize\textit{f)} Frame 110}
    \end{minipage}
    
    \caption{Qualitative \gls{mot} results on \gls{bdd} validation set for our proposed transformer tracking architecture and pre-trained on a sequential ordering task. The network tracks the pedestrian reliably over changing body orientations and handles the changing shape of the green car.}
    \label{fig:qualitative_example_mot}
\end{figure*}

\begin{table*}
\footnotesize
\centering
\caption{Comparison of different parameter initialization strategies for the downstream task of multi-object tracking on the BDD100k val set. Relative metrics are averaged across all class types.}
\label{tab:eval_mot_bdd100k_val}
\begin{tabular}{lll|lllllll} \toprule
 &  &  & \multicolumn{7}{c}{BDD100k val}   \\
Model & Detector & Initialization & mHOTA $\uparrow$ & DetA $\uparrow$ & AssA $\uparrow$ & LocA $\uparrow$ & mMOTA $\uparrow$ & IDF1$\uparrow$ & IDSw. $\downarrow$ \\ \midrule
QD Track & Faster R-CNN & BDD100k & 38.3 & 38.4 & \textbf{39.6} & 68.5 & 29.3 & \textbf{49.7} & \textbf{90590} \\
QD Track & Sparse R-CNN & BDD100k & 37.2 & 38.2 & 32.3 & 69.5 & 28.7 & 43.9 & 94688 \\
TempO & Sparse R-CNN & TempO & 36.6 & 35.2 & 30.1 & 66.1 & 26.3 & 40.1 & 108755 \\
TempO & Sparse R-CNN & TempO+BDD100k & \textbf{39.2} & \textbf{39.6} & 38.7 & \textbf{69.6} & \textbf{31.2} & 46.8 & 91083 \\ \bottomrule
\end{tabular}
\end{table*}

\begin{table*}
\footnotesize
\centering
\caption{Ablation experiments on downstream tasks for various TempO pre-training settings \change{for six epochs on the BDD100k dataset}, using a Sparse R-CNN model with a ResNet-50 backbone and 100 proposals per frame. Relative metrics are averaged across all class types.}
\label{tab:ablation}
\begin{tabular}{cccc|cc|p{0.35cm}p{0.35cm}p{0.35cm}p{0.35cm}p{0.35cm}p{0.35cm}|llll}
\toprule
\multicolumn{4}{c}{Pretext setting} & \multicolumn{2}{c}{\change{Pre-train times}} & \multicolumn{6}{c}{BDD100k val object detection} & \multicolumn{4}{c}{BDD100k val MOT} \\
$N_{seq}$ & $L_{\{enc, dec\}}$ & $f_{sim}$ & Augment. & Iter/s & GPUh & AP & AP50 & AP75 & APs & APm & APl & HOTA ↑ & MOTA ↑ & IDF1 ↑ & IDSw. ↓ \\ \midrule
4 & 2 & AvgPool & -  & 12.1 & 551 & 31.0 & 56.5 & 29.0 & 15.2 & 34.8 & 51.5 & 33.6 & 25.5 & 37.5 & 102261 \\
6 & 2 & AvgPool & -  & 9.2 & 478 & 31.2 & 56.8 & 29.3 & 15.3 & 35.0 & 51.4 & 34.9 & 25.0 & 36.4 & 101723 \\
8 & 2 & AvgPool & -  & 8.2 & 407 & \textbf{31.4} & \textbf{57.3} & \textbf{29.5} & \textbf{15.7} & \textbf{35.2} & \textbf{51.6} & 36.6 & 26.3 & 40.5 & 90388\\
8 & 1 & AvgPool & -  & 8.7 & 383 & 29.1 & 54.1 & 26.9 & 14.2 & 33.1 & 48.0 & 35.1 & 24.4 & 36.1 & 92529 \\
8 & 4 & AvgPool & -  & 7.6 & 439 & 30.9 & 56.5 & 28.7 & 15.2 & 34.8 & 50.7 & \textbf{37.2} & \textbf{27.2} & \textbf{42.9} & \textbf{89083} \\
8 & 2 & AvgPool & P  & 8.2 & 408 & 31.3 & 56.9 & 29.3 & 15.4 & 35.1 & 51.2 & 35.5 & 15.8 & 41.5 & 92053\\
\bottomrule
\end{tabular}
\end{table*}

We evaluate our TempO pre-trained models on the \gls{mot} downstream task using CLEAR and \gls{hota}~\cite{luiten2020hota} metrics. We build on QDTrack~\cite{pang2021quasi} to extend our model into a multi-object tracker, as it associates detected objects solely in feature space. As a baseline, we trained the QDTrack~\cite{pang2021quasi} method using the Faster R-CNN and Sparse R-CNN model as a detector, which was each pre-trained using annotations of the \gls{bdd} detection dataset.

\tabref{tab:eval_mot_bdd100k_val} shows the performance on the \gls{bdd} \gls{mot} val dataset for fine-tuning with various parameter initialization strategies on the \gls{bdd} \gls{mot} training annotations for 4 epochs. 
As expected for only four fine-tuning epochs, the initialization with TempO pre-training only results in lower overall tracking accuracy compared to a strictly supervised pre-training on the \gls{bdd} detection set. 
However, we observe that combined TempO pre-training followed by fine-tuning as an object detector achieves the highest tracking accuracy in both the mHOTA and mMOTA scores averaged over all classes, outperforming the baseline QDTrack using a Faster R-CNN detector by $+0.9\%$ in the \gls{hota} score. 
Interestingly, using Sparse R-CNN as a base detector achieves consistently lower association accuracy and IDF1 scores than Faster R-CNN. 
This results from the architectural difference between the QDTrack models, whereby the Faster R-CNN variant uses a RoI head while the Sparse R-CNN variant relies on two Dynamic Head modules to compute tracking features. For comparison among the Sparse R-CNN QDTrack variants, we observe that self-supervised TempO pre-training increases the IDF1 score compared to supervised pre-training of the detector alone.\looseness=-1

In \secref{sec:egomotion} of the supplementary material, we present an experiment to study how TempO pre-training influences the performance of the tracking model with respect to the ego-motion of the camera. The results show that the TempO-initialized model demonstrates superior tracking performance due to more consistent tracking accuracy across various ego-motion speeds.

\figref{fig:qualitative_example_mot} shows an example tracking sequence using the tracking architecture described in \secref{sec:downstream_networks}. The network reliably tracks the car reliably, even under heavy occlusion and changing shape from opening the car door. The pedestrian entering the car is successfully tracked over the front, side, and rearview and under partial occlusion when stepping into the backseat.

{\parskip=3pt
\noindent\textit{MOT17 Results}:
We further evaluate our TempO pre-training strategy on the popular MOT17 benchmark. We compare model initialization by our proposed TempO pre-training against supervised training on the COCO 2017 dataset, as well as unsupervised pre-training on the Crowdhuman~\cite{shao2018crowdhuman} dataset by an RPD pretext task. 
We follow the training setting described in QDTrack and fine-tune our models for 12 epochs on mixed CrowdHuman and MOT17 train sets. The result in \tabref{tab:eval_mot17} indicates that TempO pre-trained Sparse R-CNN model outperforms both supervised and unsupervised initialization strategies by  $+1.4\%$ in the \gls{hota} score and $+2.3\%$ in MOTA, which results from increased detection accuracy DetA and fewer ID switches.}

\begin{table*}
\footnotesize
\centering
\caption{Comparison of parameter initialization strategies for multi-object tracking on MOT17 test set. All detectors are fine-tuned on CrowdHuman after pre-training.}
\label{tab:eval_mot17}
\begin{tabular}{ll|ll|lllllllll} \toprule
\multicolumn{2}{c}{Model} & \multicolumn{2}{c}{\change{Pre-train}} & \multicolumn{9}{c}{MOT17 test set}                                                                                                               \\ 
Tracker   & Detector      & Dataset        & Task         & MOTA↑         & IDF1↑         & HOTA↑         & DetA          & AssA          & LocA          & FP↓            & FN↓             & IDSw.           \\ \midrule
QDTrack   & Faster R-CNN  & COCO           & Det         & 68.7          & \textbf{66.3} & 53.9          & -             & -             & -             & 26589          & 146643          & 3378          \\
QDTrack   & Sparse R-CNN  & COCO           & Det         & 69.5          & 63.4          & 52.9          & 56.2          & 49.1          & 82.2          & 21963          & 147291          & 3305          \\
QDTrack   & Sparse R-CNN  & MOT17          & RPD          & 70.8          & 65.9          & 52.1          & 56.7          & 48.4          & 80.7          & 42855          & 117402          & 4563          \\ \toprule
QDTrack   & DDETR         & MOT17          & TempO        & 72.1          & 63.9          & 53.2          & 57.9          & 49.1          & 82.4          & 18513          & 135687          & 3180          \\
QDTrack   & Sparse R-CNN  & MOT17          & TempO        & \textbf{72.8} & 65.9          & \textbf{54.3} & \textbf{58.5} & \textbf{50.3} & \textbf{82.5} & \textbf{17646} & \textbf{133059} & \textbf{3093} \\ \bottomrule
\end{tabular}
\end{table*}

\subsection{Ablation Study}
\label{sec:ablation_study}

\tabref{tab:ablation} presents the benchmarking results for object detection and \gls{mot} downstream tasks on the \gls{bdd} dataset for models pre-trained on various TempO configurations. In particular, we ablate over the sequence length, the attention hierarchies of the multi-frame transformer encoder, and the choice of temporally-varying frame augmentations.
    
{\parskip=3pt 
\noindent\textit{Sequence Length}:
By varying the number of subsequent frames from 4 to 8, the bounding box AP increases from 31.0\% to 31.4\%, demonstrating that a longer temporal context allows the model to learn more distinctive object attributes to detect object types reliably. For \gls{mot} performance, the gain from observing longer sequences and, therefore, more framewise comparisons are as high as $+3\%$ in the \gls{hota} score.} 
    
{\parskip=3pt
\noindent\textit{Hierarchical Attention}:
Another vital design aspect is the size of the multi-frame network, and especially the hierarchy of associations that can be increased by stacking multiple encoder layers. The ablation study shows that this hyperparameter has a big impact on the performance of the downstream task compared to the sequence length. Surprisingly, the complexity is saturated at two encoder layers, while a higher number of layers decreases performance, especially on single-frame object detection tasks. 
We initially hypothesized that the multi-frame model could be incapable of generalizing across all the dynamic interactions that inform about temporal sequences in traffic scenes. However, the results show that a lighter multi-frame head loads more semantic reasoning onto the single-frame model, thereby learning more expressive features. Moreover, the damped performance can result from the slower convergence due to the high parallelism in multi-head attention modules, such that longer pre-training schedules can be required. }
    
{\parskip=3pt
\noindent\textit{Spatial and Photometric Augmentations}:
In \tabref{tab:ablation}, we evaluate spatial and appearance-based augmentations of the input sequence. This enforces the network to learn object representations that are invariant to the global image location or lightning effects. Interestingly, photometric augmentations during pre-training resulted in lower performance on the downstream task, reducing the object detection performance by $-0.1\%$ in $mAP$ and tracking performance by $-1.4\%$ in the \gls{hota} score for identical TempO settings. This shows that the network exploits internal appearance consistency assumptions. Spatial augmentations such as random cropping negatively affect the pre-training, which indicates that the network relies on consistent spatial cues to solve the temporal ordering task.}

\subsection{Frame Retrieval Experiments}
\label{sec:frame_retrivial}
We further perform frame retrieval experiments on the UCF101 dataset~\cite{soomro2012ucf101} for a direct comparison with related image-level self-supervised methods of jig-saw ordering~\cite{ahsan2019video} or temporal ordering~\cite{OPN,kong2020ccl}.
We trained our models for 100 epochs on the UCF101 videos using the \textit{TempO} pretext task as described in \secref{sec:ordering_task}. To compute the closeness between two frames, we performed an average pooling of pairwise association scores computed from the proposal feature vectors of two frames. We classified samples as correct if a vector belonging to the same action class was within the $k$ nearest neighbors.

\begin{table}[t]
\footnotesize
\setlength{\tabcolsep}{5pt}
\centering
\caption{Frame retrieval results on UCF101 dataset. }
\label{tab:frame_retrieval_ucf101}
\begin{tabular}{llllll}
\toprule
Method & Model & Top-1 & Top-5 & Top-10 & Top-20 \\ \midrule
MSE & - & 13.1 & 20.2 & 23.4 & 28.6 \\
JigSaw~\cite{ahsan2019video} & 3D CNN & 19.7 & 28.5 & 33.5 & 40.0 \\
OPN~\cite{OPN} & 3D CNN & 19.9 & 28.7 & 34.0 & 40.6 \\
CCL~\cite{kong2020ccl} & 3D ResNet & 32.7 & 42.5 & 50.8 & 61.2 \\
\midrule
\multirow{3}{*}{TempO} & Faster R-CNN & 34.9 & 46.1 & 53.6 & 58.9 \\
 & Sparse R-CNN & \textbf{35.6} & \textbf{49.5} & \textbf{58.2} & \textbf{68.3} \\
 & DDETR & 33.1 & 18.4 & 56.4 & 65.9 \\ \bottomrule
\end{tabular}
\end{table}

We evaluated our method against the baseline performances reported by \textit{Kong et al.}~\cite{kong2020ccl}. In \tabref{tab:frame_retrieval_ucf101}, we present the retrieval accuracy for $k=1$ to $k=20$. The results show that our TempO pre-trained embeddings exhibit strong consistency across videos belonging to the same action class. Notably, our approach outperforms other frame-level pre-training strategies, including the cycle-consistency pretext (CCL) task, and improves the Top-1 accuracy by 3.9\%. To give an idea of how our approach works, we provide examples of misclassified frames and their top three most similar class representatives in \figref{fig:bdd100k_retrieval_errors}. Our observations show that TempO representations focus primarily on the similarity of scene attributes, such as the number and size of objects or camera perspective. In particular, the example at the bottom of \figref{fig:bdd100k_retrieval_errors} indicates that the learned representation contains more tracking features on foreground objects, given the large variety of backgrounds.
\begin{figure}[t]
    \centering
    \footnotesize
    \begin{tabular}{c|cccc} 
        Query image & Top 1 & Top 2 & True class\\
        \\[0.5em]
        \includegraphics[scale=0.15]{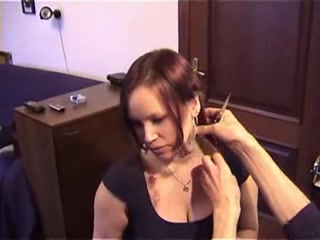} &
        \includegraphics[scale=0.15]{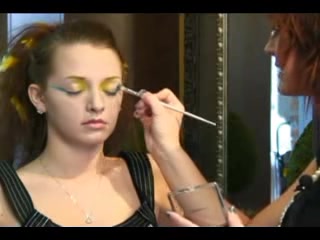} & 
        \includegraphics[scale=0.15]{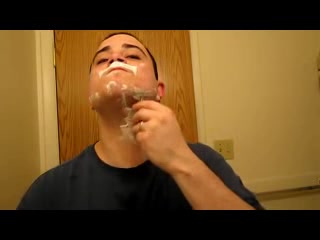} &
        \includegraphics[scale=0.15]{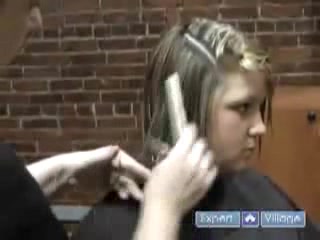} 
        \\
                 &
        \begin{tabular}{@{}c@{}}Apply eye makup: \\ 0.27\end{tabular}  & 
        \begin{tabular}{@{}c@{}}Shaving beard: \\ 0.22\end{tabular}  &
        \begin{tabular}{@{}c@{}}Hair cutting \\ \end{tabular}  
        \\[1em]
        \includegraphics[scale=0.15]{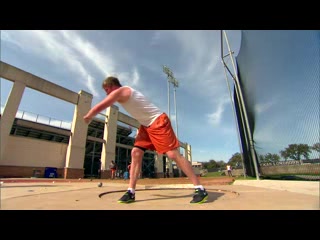} &
        \includegraphics[scale=0.15]{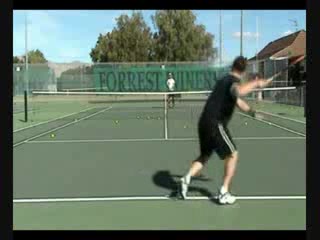} & 
        \includegraphics[scale=0.15]{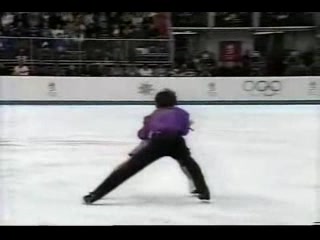} &
        \includegraphics[scale=0.15]{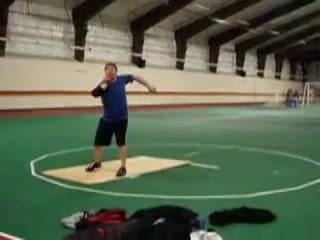} 
        \\
                 &
        \begin{tabular}{@{}c@{}}Tennis swing: \\ 0.21\end{tabular}  & 
        \begin{tabular}{@{}c@{}}Ice dancing: \\ 0.17\end{tabular} &
        \begin{tabular}{@{}c@{}}Shotput \\ \end{tabular}    
    \end{tabular}
    \caption{Frame retrieval demonstrations of misclassifications in the top-20 setting. Scores are normalized similarities of the top-20 nearest neighbors. Retrieved frames resemble scene attributes, for instance, the number and size of foreground objects or camera perspective, while backgrounds vary widely.}
    \label{fig:bdd100k_retrieval_errors}
\end{figure}

\subsection{Discussion of Limitations}

Our analysis and the general self-supervised video feature learning field focus on relatively short clips of $<2s$. Many human or driving actions, however, extend over longer periods, and our ablation study shows that longer sequences can benefit the pre-trained models. 
Secondly, we perform most of our evaluation on driving sequences, where the camera moves smoothly in a dynamic environment. 
Even though we also evaluate the MOT17 dataset, which is human-centric and from predominantly static cameras, future work could evaluate how TempO pre-training behaves on a mixture of domains or highly repetitive videos, \eg~from an indoor service robot. 

    \section{Conclusion}
    In this work, we proposed a \gls{ssl} pretext task based on the temporal ordering of video frames to pre-train almost all parameters of object detection and \gls{mot} networks.
Models initialized with TempO pre-trained weights demonstrated a speed-up in convergence and superior performance on object detection and multi-object tracking downstream tasks compared to other self-supervised as well as supervised initialization strategies. The qualitative results also show how TempO pre-training helps suppress ghost detection and recognize dark objects at night from a semantic context. In our multi-object tracking experiments, TempO pre-training improves the tracking accuracy by $+1.4\%$ in the HOTA score while using a pre-trained object detector and initializing the tracking branch with TempO pre-trained weights. We further evaluated the learned representations without fine-tuning for a frame retrieval task, where we found that representations across videos are more consistent for similar action classes compared to related works in self-supervised pretraining from video. We find that the temporal ordering pretext task can boost performance compared to single-frame or supervised pre-training strategies in instance-level perception systems. 

{\footnotesize
\bibliographystyle{IEEEtran}
\bibliography{references_ssl}  
}



\clearpage

\begin{strip}
\begin{center}
\vspace{-5ex}
\textbf{\LARGE \bf
Self-Supervised Representation Learning from Temporal Ordering of\\Automated Driving Sequences} \\
\vspace{3ex}

\Large{\bf- Supplementary Material -}\\
\vspace{0.4cm}

\normalsize{Christopher Lang$^{1,2}$, Alexander Braun$^{2}$, Lars Schillingmann$^{2}$, Karsten Haug$^{2}$, and Abhinav Valada$^{1}$}
\end{center}
\end{strip}

\setcounter{section}{0}
\setcounter{equation}{0}
\setcounter{figure}{0}
\setcounter{table}{0}
\makeatletter

\renewcommand{\thesection}{S.\arabic{section}}
\renewcommand{\thesubsection}{S.\arabic{subsection}}
\renewcommand{\thetable}{S.\arabic{table}}
\renewcommand{\thefigure}{S.\arabic{figure}}

\renewcommand{\theHsection}{S.\arabic{section}}
\renewcommand{\theHsubsection}{S.\arabic{subsection}}
\renewcommand{\theHtable}{S.\arabic{table}}
\renewcommand{\theHfigure}{S.\arabic{figure}}

\let\thefootnote\relax\footnote{$^{1}$Robert Bosch GmbH, Stuttgart, Germany\\
$^{2}$Department of Computer Science, University of Freiburg, Germany}

\normalsize
In this supplementary material, we first present details on hyperparameter tuning and computational requirements in \secref{sec:implementation}, followed by additional experimental results on the convergence behavior for varying initialization strategies in \secref{sec:convergence}, and an evaluation of the tracking performance with respect to ego-motion of the camera in \secref{sec:egomotion}.

\begin{figure*}[t]
    \footnotesize
    \centering
    \begin{minipage}[t]{0.49\textwidth}
        \centering
        \begin{minipage}{0.48\textwidth}
            \includegraphics[trim={6cm 0 0 0},clip,width=\textwidth]{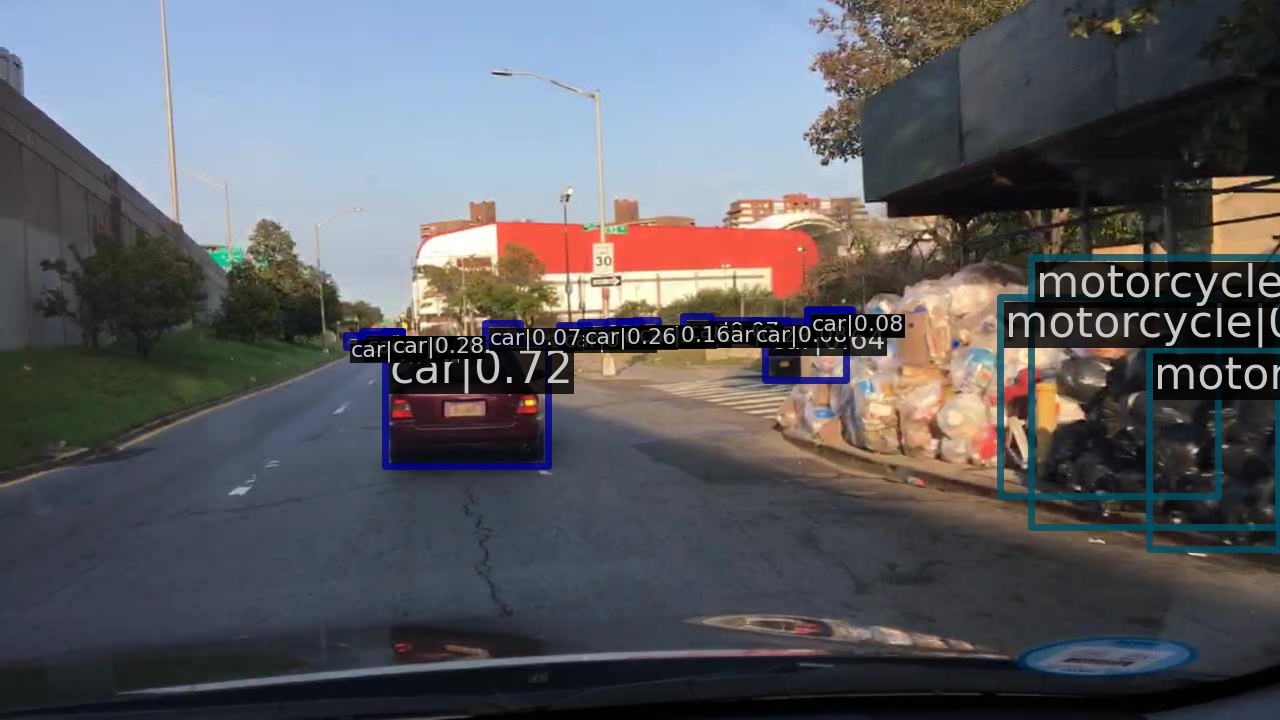}
        \end{minipage}%
        \hspace{0.01\textwidth}%
        \begin{minipage}{0.48\textwidth}\centering
            \includegraphics[trim={0 0 6cm 0},clip,width=\textwidth]{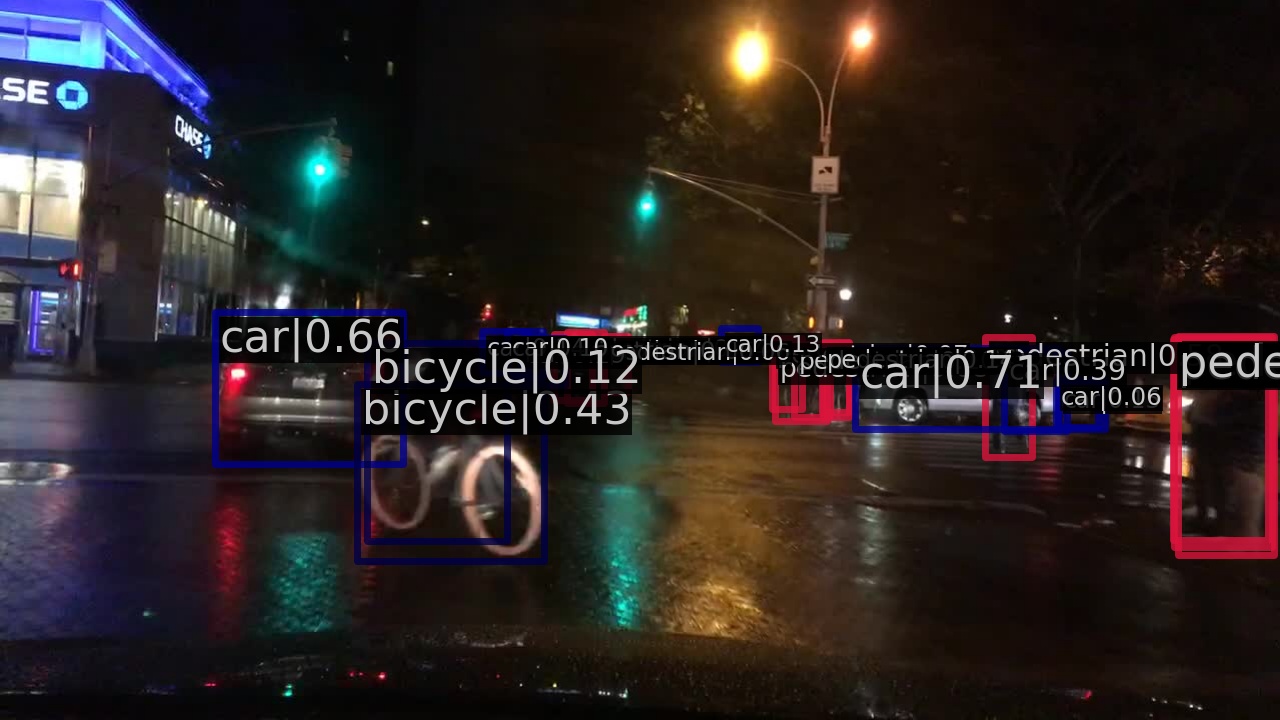}
        \end{minipage}\\[1mm]{a) Initialized with ImageNet weights and trained for 6 epochs on BDD100k.}
    \end{minipage}
    \hfill
    \begin{minipage}[t]{0.49\textwidth}
        \centering
        \begin{minipage}{0.48\textwidth}
            \includegraphics[trim={6cm 0 0 0},clip,width=\textwidth]{figures/detection_results/tempo/06_epochs/b3a6f586-23d13c49.jpg}
        \end{minipage}%
        \hspace{0.01\textwidth}%
        \begin{minipage}{0.48\textwidth}\centering
            \includegraphics[trim={0 0 6cm 0},clip,width=\textwidth]{figures/detection_results/tempo/06_epochs/bca10f4f-2deaf782.jpg}
        \end{minipage}\\[1mm]{b) TempO pre-trained and fine-tuned for 6 epochs on BDD100k.}
    \end{minipage}
\\
    \vspace{5pt}
    \begin{minipage}[t]{0.49\textwidth}
        \centering
        \begin{minipage}{0.48\textwidth}
            \includegraphics[trim={6cm 0 0 0},clip,width=\textwidth]{figures/detection_results/baseline/12_epochs/b3a6f586-23d13c49.jpg}
        \end{minipage}%
        \hspace{0.01\textwidth}%
        \begin{minipage}{0.48\textwidth}\centering
            \includegraphics[trim={0 0 6cm 0},clip,width=\textwidth]{figures/detection_results/baseline/12_epochs/bca10f4f-2deaf782.jpg}
        \end{minipage}\\[1mm]{c) Initialized with ImageNet weights and trained for 12 epochs on BDD100k.}
    \end{minipage}
    \hfill
    \begin{minipage}[t]{0.49\textwidth}
        \centering
        \begin{minipage}{0.48\textwidth}
            \includegraphics[trim={6cm 0 0 0},clip,width=\textwidth]{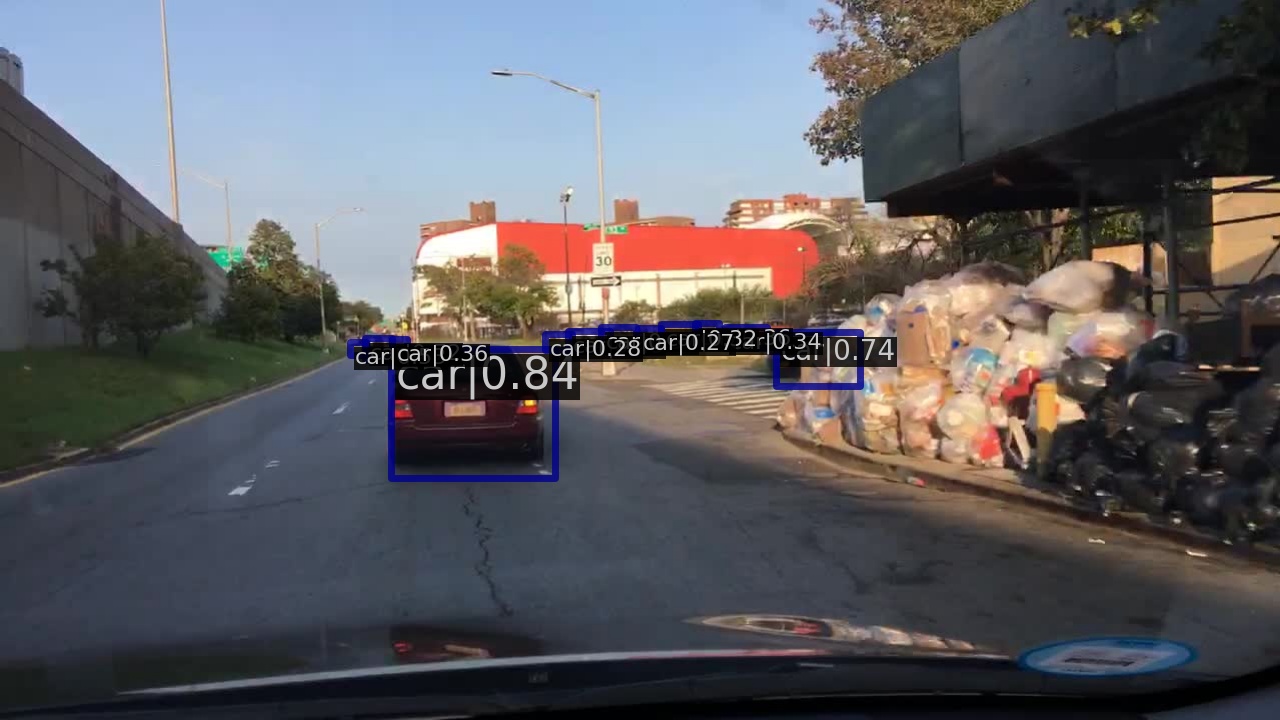}
        \end{minipage}%
        \hspace{0.01\textwidth}%
        \begin{minipage}{0.48\textwidth}\centering
            \includegraphics[trim={0 0 6cm 0},clip,width=\textwidth]{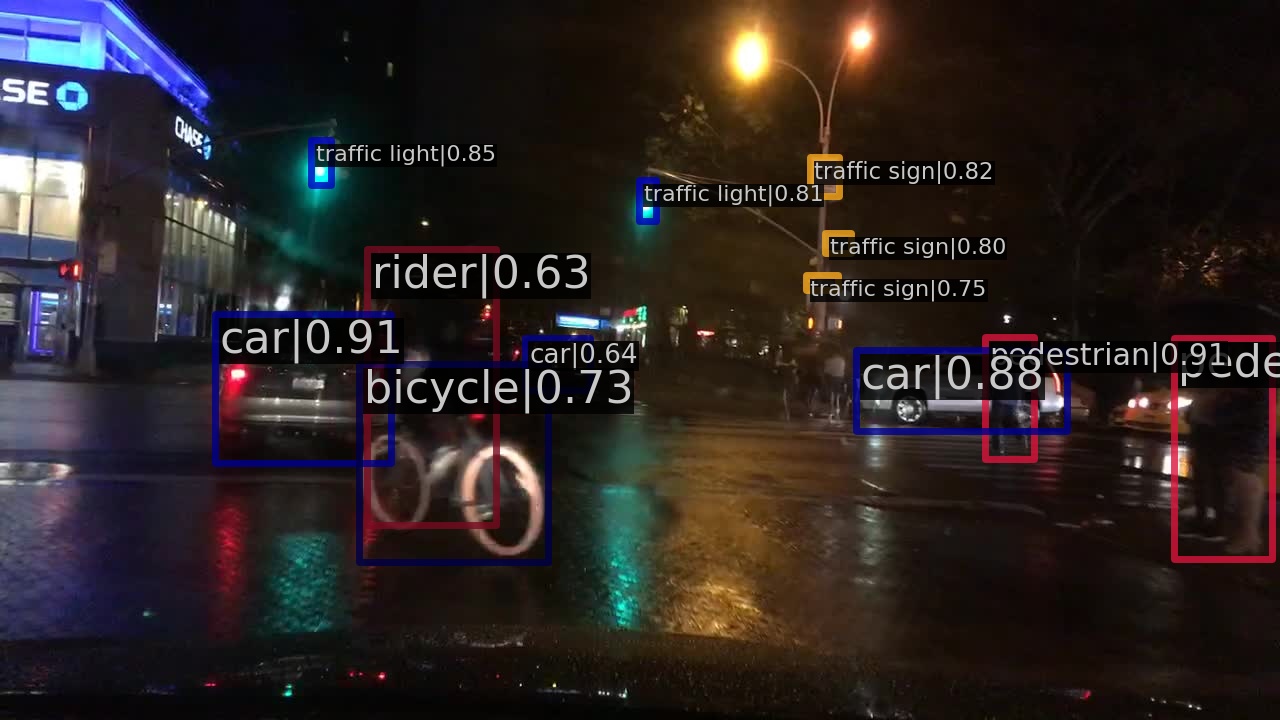}
        \end{minipage}\\[1mm]{d) TempO pre-trained, fine-tuned for 12 epochs on BDD100k.}
    \end{minipage}
    \caption{Comparison of qualitative object detection results on the BDD100k val set using the Sparse R-CNN detector with standard (a,c) and TempO (b,d) parameter initialization strategies. 
    Observe that the TempO pre-trained detector avoids a ghost detection of a motorcycle within the garbage bags and detects the poorly lit rider on top of the moving bicycle.}
    \label{fig:qualitative_example_od_rebuttal}
\end{figure*}


\section{Details of Training and Hyperparameter Tuning}
\label{sec:implementation}

\subsection{Hyperparameter Tuning}

In this section, we provide additional details about our design choices and hyperparameter tuning processes.

\noindent \paragraph{Network architecture} We selected the QDTrack multi-object tracking logic as it considers re-identification feature-based tracking, and an official baseline on the BDD100k dataset is provided for this method. We chose the number of dynamic heads during pre-training by a grid search by fine-tuning on the BDD100k MOT benchmark.

\noindent
\paragraph{Training hyperparameters} The image size and image pre-processing were adapted from the baseline configuration of the Sparse R-CNN detector model for the BDD100k dataset~\citeSuppl{suppl_yu2020bdd100k}. We adapted the training configuration for finetuning according to the 1x schedule of the benchmark. For the pre-training configuration, we chose a batch size of 8 to fit a maximum sequence of 8 frames into the GPU-RAM of four NVIDIA V100 GPUs we used for our experiments.
The number of pre-training epochs was set to 6 by observing the rate-of-change of the loss value on the BDD100k training set, summarized in \tabref{tab:eval_pretrain_epochs} in \secref{sec:convergence}.
We used the ranking loss value (\equref{eq:tempo}) to monitor performance during pre-training and used the video sequences of the validation set to detect overfitting. However, with our initial hyperparameter setting, training and validation set loss declined steadily for six training epochs on the BDD100k pre-training dataset without indicating overfitting.

\subsection{Computational Resource Requirements}

\begin{table*}[ht]
\scriptsize
\caption{Overview of computational resources and training times for various \textit{TempO} configurations on the BDD100k dataset. Floating point operations (FLOPs) are measured for the forward pass of the Sparse R-CNN model and the loss computation.}
\label{tab:computational_complexity}
\begin{tabular}{ll|lll|ll|ll|lll|lll|ll} \toprule
\multicolumn{2}{l|}{TempO}                         & \multicolumn{10}{c|}{Computational requirements}                                                                                                              & \multicolumn{5}{c}{BDD100k dataset}                                                    \\
\multicolumn{2}{l|}{Config}                              & \multicolumn{3}{c|}{\cellcolor{CornflowerBlue!25}Single-frame network}                  & \multicolumn{7}{c|}{\cellcolor{Thistle!25}Multi-frame network}                                                                           & \multicolumn{3}{c|}{Pre-training} & \multicolumn{2}{c}{Downstream-Task}             \\
\multicolumn{2}{l|}{}                              &  \cellcolor{CornflowerBlue!25}             & \cellcolor{CornflowerBlue!25}           & \cellcolor{CornflowerBlue!25}                 & \multicolumn{2}{l|}{Transfromer encoder} & \multicolumn{2}{l|}{Additive attention} & \multicolumn{3}{l|}{\cellcolor{Thistle!25}Total}     &             &            &             &                &                      \\
$N$                         & $L$                  & \cellcolor{CornflowerBlue!25}Params        & \cellcolor{CornflowerBlue!25}FLOPs      & \cellcolor{CornflowerBlue!25}FLOPs/N          & Params           & FLOPs        & Params              & FLOPs            & \cellcolor{Thistle!25}Params   & \cellcolor{Thistle!25}FLOPs & \cellcolor{Thistle!25}FLOPs/N     & \# D        & D/s        & GPUh        &  mAP           &   mHOTA              \\ \midrule
4                           & 2                    & \cellcolor{CornflowerBlue!25}53.3M         & \cellcolor{CornflowerBlue!25}242G       & \cellcolor{CornflowerBlue!25}60.5G            & 2.11M            & 1.17G        & 132k                & 62M            & \cellcolor{Thistle!25}2.24M    & \cellcolor{Thistle!25}1.23G & \cellcolor{Thistle!25}0.308G      & 4M          & 12.1       & 551.0       & 31.0           & 33.6                 \\
6                           & 2                    & \cellcolor{CornflowerBlue!25}53.3M         & \cellcolor{CornflowerBlue!25}363G       & \cellcolor{CornflowerBlue!25}60.5G            & 2.11M            & 2.00G        & 132k                & 130M             & \cellcolor{Thistle!25}2.24M    & \cellcolor{Thistle!25}2.13G & \cellcolor{Thistle!25}0.355G      & 2.64M       & 9.2        & 478.3       & \underline{31.2}     & 34.9           \\
8                           & 2                    & \cellcolor{CornflowerBlue!25}53.3M         & \cellcolor{CornflowerBlue!25}484G       & \cellcolor{CornflowerBlue!25}60.5G            & 2.11M            & 3.00G        & 132k                & 217M             & \cellcolor{Thistle!25}2.24M    & \cellcolor{Thistle!25}3.21G & \cellcolor{Thistle!25}0.402G      & 2M          & 8.7        & 406.5       & \textbf{31.4}  & \underline{36.6}     \\
8                           & 1                    & \cellcolor{CornflowerBlue!25}53.3M         & \cellcolor{CornflowerBlue!25}484G       & \cellcolor{CornflowerBlue!25}60.5G            & 1.05M            & 1.50G        & 132k                & 217M             & \cellcolor{Thistle!25}1.18M    & \cellcolor{Thistle!25}1.72G & \cellcolor{Thistle!25}0.214G      & 2M          & 8.2        & 383.14       & 29.1           & 35.1                 \\
8                           & 4                    & \cellcolor{CornflowerBlue!25}53.3M         & \cellcolor{CornflowerBlue!25}484G       & \cellcolor{CornflowerBlue!25}60.5G            & 4.21M            & 5.99G        & 132k                & 217M             & \cellcolor{Thistle!25}4.34M    & \cellcolor{Thistle!25}6.21G & \cellcolor{Thistle!25}0.776G      & 2M          & 7.6        & 438.6         & 30.9           & \textbf{37.2}        \\ \bottomrule
\end{tabular} 
\end{table*}

In \tabref{tab:computational_complexity}, we summarize the computational requirements of the ablation experiments from \tabref{tab:ablation} in the main paper. 
We report the number of parameters and the floating point operations (FLOPs) for the single-frame network in blue and the multi-frame network in purple. 
\textit{N} denotes the sequence length in frames, and \textit{L} denotes the number of transformer layers used for predicting the temporal order of frames in the TempO loss computation. 
We pre-trained the networks for six epochs on the BDD100k train set, whereby the number of training samples per epoch \textit{D} varies with the sequence length. We also report the resulting throughput of training samples per second \textit{D/s}, that we achieved on our training setup consisting of 4x Nvidia V100 GPUs and a batch size of 8.
As the single-frame network, we use the Sparse R-CNN~\citeSuppl{suppl_peize2020sparse} that is described in \secref{sec:singe_frame_network}, which processes each frame separately to extract a set of 100 proposal feature vectors per frame.
The multi-frame network consists of a transformer encoder and an additive attention layer, as depicted in \figref{fig:loss_architecture}, and jointly processes the proposal feature vectors of all frames in the sequence to compute the proposed TempO loss.

Due to the transformer-based multi-frame model architecture, the number of trainable parameters is independent of the sequence length $N$. For a multi-frame configuration using two transformer layers, the ratio of parameters in the single-frame network to the multi-frame network is approximately $25:1$. 
The FLOPs in the multi-task network scale less than quadratically with the sequence length due to our formulation of the ordering task using frame-transition probabilities.

We summarize the training times for pre-training on the BDD100k dataset for six epochs in the rightmost section of \tabref{tab:computational_complexity}. 
Moreover, we provide the downstream-task performances of the pre-trained model after fine-tuning them on the object detection and multi-object tracking downstream tasks from \tabref{tab:ablation} of the main paper.
In general, training with data samples of longer sequence lengths resulted in shorter training durations per epoch.
However, this results from our design choice to generate the pre-training set by extracting distinct sequences at five frames per second from the videos of the BDD100k train set, ensuring that each frame appeared only once per epoch. The training throughput measured in data samples per second ($D/s$) decreases with increasing the sequence length per data sample.
Although this results in fewer parameter updates during pre-training, we found that the parameter initialization pre-trained on a sequence length of eight frames resulted in the best detection and tracking performance.

\section{Convergence Experiments}
\label{sec:convergence}

\subsection{Effect on Finetuning Convergence}

In \tabref{tab:ablation_det_convergence}, we compare the performance of detectors over the number of finetuning epochs.
The detector parameters were either initialized with TempO pre-trained weights or model parameters trained on the COCO 2017 dataset.
The TempO pre-training used a sequence length of 8 frames, two layers in the multi-frame network, and AvgPool.
All the detectors were fine-tuned with a batch size of 8 for 12 epochs on the \acrshort{bdd} train split with the training settings described in \secref{sec:datasets} of the main paper.

\begin{table}[ht]
\scriptsize
\centering
\caption{Comparison of parameter initialization strategies on the BDD100k object detection task for an increasing number of fine-tuning epochs. }
\label{tab:ablation_det_convergence}
\begin{tabular}{p{1.3cm}p{0.75cm}p{0.5cm}|p{0.35cm}p{0.35cm}p{0.35cm}p{0.35cm}p{0.35cm}p{0.35cm}} \toprule
Model & Pretrain & Epoch &  \multicolumn{6}{c}{BDD100k val object detection} \\
& & & AP & AP50 & AP75 & APs & APm & APl \\ \midrule
\multirow{8}{*}{SparseRCNN} & \multirow{4}{*}{COCO} & 2 & 25.3 & 47.6 & 23.1 & 12.5 & 28.4 & 40.3 \\
 &  & 4 & 27.7 & 51.0 & 25.8 & 13.7 & 31.2 & 44.5 \\
 &  & 6 & 27.5 & 49.9 & 25.8 & 12.6 & 30.5 & 47.8 \\ 
 &  & 12 & 30.7 & \underline{55.8} & \underline{28.9} & 15.2 & 34.3 & \underline{50.8} \\ \cmidrule{2-9}
 & \multirow{4}{*}{TempO} & 2 & 23.4 & 46.0 & 20.8 & 11.0 & 27.0 & 36.2 \\
 &  & 4 & 28.6 & 52.5 & 26.9 & 13.6 & 32.4 & 48.3 \\
 &  & 6 & \underline{30.7} & 55.5 & 28.6 & \textbf{15.4} & \underline{34.6} & 49.6 \\
 &  & 12 & \textbf{31.4} & \textbf{57.2} & \textbf{29.3} & \underline{15.3} & \textbf{35.2} & \textbf{52.4} \\  \midrule
 \multirow{8}{*}{DDETR} & \multirow{4}{*}{COCO} & 2 & 9.3 & 20.9 & 6.9 & 4.5 & 11.6 & 17.3 \\
 &  & 4 & 16.8 & 34.7 & 14 & 7.6 & 19.8 & 30.6 \\
 &  & 6 & 28.3 & 52.2 & 26.1 & 11.2 & 32.1 & 53.6 \\
 &  & 12 & 30.2 & 56.0 & 27.6 & 14.2 & 34.0 & 51.3 \\ \cmidrule{2-9}
 & \multirow{4}{*}{TempO} & 2 & 13.9 & 29.9 & 11.3 & 6.1 & 16.6 & 26.6 \\
 &  & 4 & 30.6 & 55.9 & 28.3 & 12.7 & 34.3 & 55.0 \\
 &  & 6 & \underline{31.3} & \underline{57.9} & \underline{28.9} & \underline{15.2} & \underline{34.9} & \underline{55.3} \\
 &  & 12 & \textbf{32.5} & \textbf{59.2} & \textbf{30.4} & \textbf{15.7} & \textbf{36.9} & \textbf{55.3} \\ \bottomrule
\end{tabular}
\end{table}

We observe that the TempO pre-trained initialization yields the fastest convergence and achieves the highest mean average precision among different initialization strategies from 3 finetuning epochs onwards. The TempO pre-trained models, when finetuned for six epochs, produce results comparable to those achieved by the COCO pre-trained methods after 12 finetuning epochs. Moreover, they outperform the COCO pre-trained initialization by more than 0.7\% in mAP after 12 finetuning epochs. During the early finetuning epochs, TempO shows a high increase in detection performance for large and medium-sized objects, but this growth slows down after six epochs compared to COCO-initialized detectors. However, the performance achieved at this stage is already higher than that achieved by the COCO-initialized object detectors.

\subsection{Effect of Longer Finetuning Schedules}

\begin{table}[ht]
\scriptsize
\centering
\caption{Object detection performance over increasing the number of fine-tuning epochs for a Sparse R-CNN detector on the BDD100k dataset. }
\label{tab:ablation_det_convergence_20epochs}
\begin{tabular}{l|llll|llll} \toprule
      & \multicolumn{4}{c|}{TempO BDD100k init} & \multicolumn{4}{c}{Det COCO init} \\
Epoch & AP       & APs      & APm     & APl     & AP     & APs    & APm    & APl    \\ \midrule
1     & 17.3     & 7.7      & 20.3    & 29.8    & 18.6   & 8.9    & 21.0   & 33.4   \\
2     & 23.4     & 11.0     & 27.0    & 36.2    & 25.3   & 12.5   & 28.4   & 40.3   \\
4     & 28.6     & 13.6     & 32.4    & 48.3    & 27.7   & 13.7   & 31.2   & 44.5   \\
6     & 30.7     & 15.4     & 34.6    & 49.6    & 27.5   & 12.6   & 30.5   & 47.8   \\
8     & 31.2     & \underline{15.4}     & 34.9    & 51.4    & 28.7   & 13.6   & 32.3   & 48.4   \\
10    & 31.4     & \textbf{15.4}     & 35.2    & 52.1    & 30.2   & 14.6   & 33.9   & 50.4   \\
12    & 31.4    & 15.3     & 35.2    & 52.4     & 30.7   & 15.2   & 34.3   & 50.8   \\
16    & \underline{31.6}    & 15.3     & \underline{35.4}    & \underline{52.4}     & 30.7   & 15.2   & 34.3   & 51.0   \\
20    & \textbf{31.6}    & 15.3    & \textbf{35.5}    & \textbf{52.5}    & 30.9   & 15.4   & 34.4   & 51.2  \\ \bottomrule
\end{tabular}
\end{table}

In \tabref{tab:ablation_det_convergence_20epochs}, we present the object detection performance of the Sparse R-CNN detector on the BDD100k validation dataset after finetuning for 20 epochs. The parameters of the Sparse R-CNN detector were either initialized with TempO pre-trained weights or trained on the COCO dataset using bounding box annotations. The TempO pre-training uses a sequence length of 8 frames, two layers in the multi-frame network, and AvgPool.
We employed a stepwise learning rate schedule, where the base learning rate was reduced by a factor of 10 after 10 and 15 fine-tuning epochs.
By initializing the model parameters by TempO pre-trained weights, we achieved an object detection performance of $31.6$ in mAP on the \gls{bdd} val benchmark. This performance surpasses the accuracy of a comparable model with weights pre-trained on the COCO~\citeSuppl{suppl_lin2014microsoft} dataset by $+0.7$ in mAP. For both models, the mean average precision for evaluation on the BDD100k val set consistently improved with subsequent fine-tuning epochs. However, the improvement for the final eight fine-tuning epochs was as low as $0.2$ in mAP. These observations led us to finetune the models for 12 epochs on object detection benchmarks on the \gls{bdd} dataset.

\subsection{Effect of Longer Pre-training Schedules}

In \tabref{tab:eval_pretrain_epochs}, we compare performance on downstream tasks for pre-training model parameters by an increasing number of pre-training epochs. We used the TempO configuration with a sequence length of eight frames, two layers in the multi-frame network, and AvgPool. In addition to the finetuning performance, we calculated the average loss value across all training samples extracted from the 200 validation set videos of the BDD100k dataset.
\begin{table}[ht]
\scriptsize
\caption{Evaluation of TempO pre-trained parameter initializations for a Sparse R-CNN base model at varying numbers of pre-training epochs. Pre-training is done on the BDD100k dataset for the downstream task of object detection and multi-object tracking on the BDD100k val set. \textit{Loss} denotes the average loss on all samples extracted from the validation set videos.}
\label{tab:eval_pretrain_epochs}
\begin{tabular}{ll|lll|lll} \toprule
\multicolumn{2}{c|}{Pre-training} & \multicolumn{3}{c|}{BDD100k val det} & \multicolumn{3}{c}{BDD100k val MOT} \\ 
Epoch           & Loss            & mAP        & mAP50      & mAP75      & HOTA       & MOTA      & IDSw       \\ \midrule
0               & -               & 24.9       & 47.0       & 22.5       & 37.2       & 28.7      & 94688      \\ \midrule
3               & 0.211           & 29.8       & 55.2       & 27.3       & 38.0       & 30.9      & 93068      \\
6               & 0.092           & \underline{30.7}       & 55.5       & \underline{28.6}       & 39.2       & 31.2      & \underline{91083}      \\
12              & \underline{0.087}           & 30.6       & \underline{56.0}       & 28.1       & \underline{39.2}       & \underline{31.3}      & 90097      \\
18              & \textbf{0.083}           & \textbf{31.0}       & \textbf{56.5}       & \textbf{29.0}       & \textbf{39.3}       & \textbf{31.5}      & \textbf{91002}      \\ \bottomrule
\end{tabular}
\end{table}

We observe that our proposed TempO loss value declines continuously with an increasing number of pre-training epochs. Interestingly, this directly translates to downstream task performance for the multi-object tracking task, as the HOTA and MOTA metrics of the fine-tuned models increase consistently with a lower loss value on the validation set. However, the improvements are as low as $0.2$ \% in HOTA for tripling the number of pre-training epochs. For the object detection downstream task, the finetuned model pre-trained for six epochs outperforms a model pre-trained for 12 epochs by $+0.1$ in mAP. 
However, this variation lies within our observed variance due to non-linear optimization and stochastic processes in the data loader during training. 

\subsection{Qualitative insights into model convergence}
In \figref{fig:qualitative_example_od_rebuttal}, we present extended object detection results for the scenes in \figref{fig:qualitative_example_od} of the main paper, after finetuning on the object detection training set for 6 and 12 epochs, respectively.
The baseline method, depicted in \figref{fig:qualitative_example_od_rebuttal}~(a) and \figref{fig:qualitative_example_od_rebuttal}~(c) detects false positive motorcycles in the garbage stack on the left image and misses the cyclist in the right image for both training states. Overall, the trained baseline model after 12 epochs shown in \figref{fig:qualitative_example_od_rebuttal}~(c) has higher confidence scores for true positive objects and reduces duplicate detections, e.g., for the bicycle in the right image.

The TempO pre-trained method shown in \figref{fig:qualitative_example_od_rebuttal}~(b) and \figref{fig:qualitative_example_od_rebuttal}~(d) avoids the false positive motorcycle in the left image and reliably detects the rider on top of the bicycle. The model has fewer variations in the confidence scores and detection boxes when comparing the models after six epochs (shown in \figref{fig:qualitative_example_od_rebuttal}~(b)) and 12 epochs (shown in \figref{fig:qualitative_example_od_rebuttal}~(d)) of finetuning, which suggests that the model after six epochs is already in a converged state. The TempO pre-trained model after 12 epochs of fine-tuning shown in \figref{fig:qualitative_example_od_rebuttal}~(d) suppresses detections of small objects such as the distant traffic signs in the left image and the pedestrian in the right image. However, in the convergence experiments in \tabref{tab:ablation_det_convergence}, we observed that the performance for small objects ($AP_s$) only declines slightly by $-0.1\%$ for the Sparse R-CNN detector compared after 6 and 12 epochs of finetuning, while this trend cannot be observed for the DDETR detector using a TempO pre-trained parameter initialization. Therefore, we expect these observations to be within the variance of the model training with respect to stochastic processes during non-linear optimization. 

\section{Influence of TempO Pre-training on Tracking with Egomotion}
\label{sec:egomotion}

\begin{figure}[t]
    \centering
    \includegraphics[width=\linewidth]{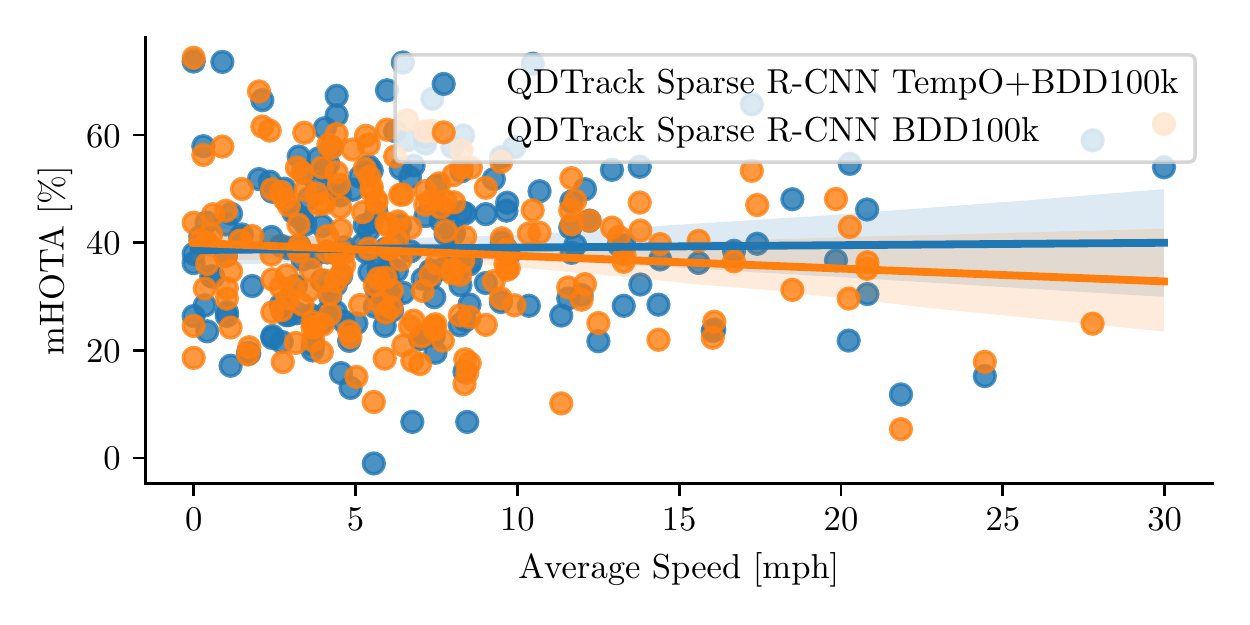}
    \caption{Per sequence HOTA vs. average velocity of ego-camera on BDD100k validation set for methods with and without TempO pretraining. 
    }
    \label{fig:hota_vs_speed}
\end{figure}

In \figref{fig:hota_vs_speed}, we compare how TempO pre-training affects the performance of the tracking model with respect to the ego-motion of the camera. Therefore, we compute regression lines from data points representing a sequence of the \gls{bdd} val set by its average ego-camera velocity and the tracking performance over the entire sequence.
The results show that the TempO pre-trained tracker performs consistently well for varying speeds of the ego-vehicle, while the fully-supervised method's performance tends to decline with faster ego-motion. We observe that the TempO pre-trained models can learn velocity-independent tracking features from the training sequences, whereas supervised methods learn associations from frame pairs.

\begin{minipage}{\columnwidth}
\footnotesize
\bibliographystyleSuppl{IEEEtran}
\bibliographySuppl{references_suppl}  
\end{minipage}

\end{document}


\title{\LARGE \bf
Self-Supervised Representation Learning from Temporal Ordering of Automated Driving Sequences\\\vspace{15px}{\Large{\bf- Supplementary Material-}}}\\

\author{Christopher Lang$^{1,2}$, Alexander Braun$^{1}$,
    Lars Schillingmann$^{1}$,  Karsten Haug$^{1}$, Abhinav Valada$^{2}$
    \thanks{$^{1}$Robert Bosch GmbH, Germany
    {\tt\small christopher.lang@de.bosch.com}}%
    \thanks{$^{2}$Department of Computer Science, University of Freiburg, Germany {\tt\small valada@cs.uni-freiburg.de}}
}

\maketitle

\begin{figure*}[t]
    \footnotesize
    \centering
    \begin{minipage}[t]{0.49\textwidth}
        \centering
        \begin{minipage}{0.48\textwidth}
            \includegraphics[trim={6cm 0 0 0},clip,width=\textwidth]{figures/detection_results/baseline/06_epochs/b3a6f586-23d13c49.jpg}
        \end{minipage}%
        \hspace{0.01\textwidth}%
        \begin{minipage}{0.48\textwidth}\centering
            \includegraphics[trim={0 0 6cm 0},clip,width=\textwidth]{figures/detection_results/baseline/06_epochs/bca10f4f-2deaf782.jpg}
        \end{minipage}\\[1mm]{a) Initialized with ImageNet weights and trained for 6 epochs on BDD100k.}
    \end{minipage}
    \hfill
    \begin{minipage}[t]{0.49\textwidth}
        \centering
        \begin{minipage}{0.48\textwidth}
            \includegraphics[trim={6cm 0 0 0},clip,width=\textwidth]{figures/detection_results/tempo/06_epochs/b3a6f586-23d13c49.jpg}
        \end{minipage}%
        \hspace{0.01\textwidth}%
        \begin{minipage}{0.48\textwidth}\centering
            \includegraphics[trim={0 0 6cm 0},clip,width=\textwidth]{figures/detection_results/tempo/06_epochs/bca10f4f-2deaf782.jpg}
        \end{minipage}\\[1mm]{b) TempO pre-trained and fine-tuned for 6 epochs on BDD100k.}
    \end{minipage}
\\
    \vspace{5pt}
    \begin{minipage}[t]{0.49\textwidth}
        \centering
        \begin{minipage}{0.48\textwidth}
            \includegraphics[trim={6cm 0 0 0},clip,width=\textwidth]{figures/detection_results/baseline/12_epochs/b3a6f586-23d13c49.jpg}
        \end{minipage}%
        \hspace{0.01\textwidth}%
        \begin{minipage}{0.48\textwidth}\centering
            \includegraphics[trim={0 0 6cm 0},clip,width=\textwidth]{figures/detection_results/baseline/12_epochs/bca10f4f-2deaf782.jpg}
        \end{minipage}\\[1mm]{c) Initialized with ImageNet weights and trained for 12 epochs on BDD100k.}
    \end{minipage}
    \hfill
    \begin{minipage}[t]{0.49\textwidth}
        \centering
        \begin{minipage}{0.48\textwidth}
            \includegraphics[trim={6cm 0 0 0},clip,width=\textwidth]{figures/detection_results/tempo/12_epochs/b3a6f586-23d13c49.jpg}
        \end{minipage}%
        \hspace{0.01\textwidth}%
        \begin{minipage}{0.48\textwidth}\centering
            \includegraphics[trim={0 0 6cm 0},clip,width=\textwidth]{figures/detection_results/tempo/12_epochs/bca10f4f-2deaf782.jpg}
        \end{minipage}\\[1mm]{d) TempO pre-trained, fine-tuned for 12 epochs on BDD100k.}
    \end{minipage}
    \caption{Comparison of qualitative object detection results on the BDD100k val set using the Sparse R-CNN detector with standard (a,c) and TempO (b,d) parameter initialization strategies. 
    Observe that the TempO pre-trained detector avoids a ghost detection of a motorcycle within the garbage bags and detects the poorly lit rider on top of the moving bicycle.}
    \label{fig:qualitative_example_od_rebuttal}
\end{figure*}


In this supplementary material, we present extended implementation details in \secref{sec:implementation}, as well as additional experimental results on the convergence behavior for varying initialization strategies in \secref{sec:convergence} and tracking performance evaluations with respect to the camera ego-motion of the camera in \secref{sec:egomotion}.


\section{Extended implementation details}
\label{sec:implementation}

\input{sections/4A_extended_implementation_details}

\subsection{Computational resources}

\begin{table*}[ht]
\scriptsize
\caption{Overview of computational resources and training times for various \textit{TempO} configurations on the BDD100k dataset. Floating point operations (FLOPs) are measured for the forward pass of the Sparse R-CNN model and the loss computation.}
\label{tab:computational_complexity}
\begin{tabular}{ll|lll|ll|ll|lll|lll|ll} \toprule
\multicolumn{2}{l|}{TempO}                         & \multicolumn{10}{c|}{Computational requirements}                                                                                                              & \multicolumn{5}{c}{BDD100k dataset}                                                    \\
\multicolumn{2}{l|}{Config}                              & \multicolumn{3}{c|}{\cellcolor{CornflowerBlue!25}Single-frame network}                  & \multicolumn{7}{c|}{\cellcolor{Thistle!25}Multi-frame network}                                                                           & \multicolumn{3}{c|}{Pre-training} & \multicolumn{2}{c}{Downstream-Task}             \\
\multicolumn{2}{l|}{}                              &  \cellcolor{CornflowerBlue!25}             & \cellcolor{CornflowerBlue!25}           & \cellcolor{CornflowerBlue!25}                 & \multicolumn{2}{l|}{Transfromer encoder} & \multicolumn{2}{l|}{Additive attention} & \multicolumn{3}{l|}{\cellcolor{Thistle!25}Total}     &             &            &             &                &                      \\
$N$                         & $L$                  & \cellcolor{CornflowerBlue!25}Params        & \cellcolor{CornflowerBlue!25}FLOPs      & \cellcolor{CornflowerBlue!25}FLOPs/N          & Params           & FLOPs        & Params              & FLOPs            & \cellcolor{Thistle!25}Params   & \cellcolor{Thistle!25}FLOPs & \cellcolor{Thistle!25}FLOPs/N     & \# D        & D/s        & GPUh        &  mAP           &   mHOTA              \\ \midrule
4                           & 2                    & \cellcolor{CornflowerBlue!25}53.3M         & \cellcolor{CornflowerBlue!25}242G       & \cellcolor{CornflowerBlue!25}60.5G            & 2.11M            & 1.17G        & 132k                & 62M            & \cellcolor{Thistle!25}2.24M    & \cellcolor{Thistle!25}1.23G & \cellcolor{Thistle!25}0.308G      & 4M          & 12.1       & 551.0       & 31.0           & 33.6                 \\
6                           & 2                    & \cellcolor{CornflowerBlue!25}53.3M         & \cellcolor{CornflowerBlue!25}363G       & \cellcolor{CornflowerBlue!25}60.5G            & 2.11M            & 2.00G        & 132k                & 130M             & \cellcolor{Thistle!25}2.24M    & \cellcolor{Thistle!25}2.13G & \cellcolor{Thistle!25}0.355G      & 2.64M       & 9.2        & 478.3       & \underline{31.2}     & 34.9           \\
8                           & 2                    & \cellcolor{CornflowerBlue!25}53.3M         & \cellcolor{CornflowerBlue!25}484G       & \cellcolor{CornflowerBlue!25}60.5G            & 2.11M            & 3.00G        & 132k                & 217M             & \cellcolor{Thistle!25}2.24M    & \cellcolor{Thistle!25}3.21G & \cellcolor{Thistle!25}0.402G      & 2M          & 8.7        & 406.5       & \textbf{31.4}  & \underline{36.6}     \\
8                           & 1                    & \cellcolor{CornflowerBlue!25}53.3M         & \cellcolor{CornflowerBlue!25}484G       & \cellcolor{CornflowerBlue!25}60.5G            & 1.05M            & 1.50G        & 132k                & 217M             & \cellcolor{Thistle!25}1.18M    & \cellcolor{Thistle!25}1.72G & \cellcolor{Thistle!25}0.214G      & 2M          & 8.2        & 383.14       & 29.1           & 35.1                 \\
8                           & 4                    & \cellcolor{CornflowerBlue!25}53.3M         & \cellcolor{CornflowerBlue!25}484G       & \cellcolor{CornflowerBlue!25}60.5G            & 4.21M            & 5.99G        & 132k                & 217M             & \cellcolor{Thistle!25}4.34M    & \cellcolor{Thistle!25}6.21G & \cellcolor{Thistle!25}0.776G      & 2M          & 7.6        & 438.6         & 30.9           & \textbf{37.2}        \\ \bottomrule
\end{tabular} 
\end{table*}

\section{Convergence Experiments}
\label{sec:convergence}

\input{sections/44_convergence}

\subsection{Qualitative insights into model convergence}
\input{sections/49_additional_convergence_experiments}





\section{Egomotion Experiments}
\label{sec:egomotion}
\input{sections/47_speed_vs_hota}

{\small
\bibliographystyle{./bibliography/ieee_fullname}
\bibliography{./bibliography/references_ssl}  
}